\documentclass[10pt,final,twocolumn]{IEEEtran}

\usepackage{xcolor}

\ifCLASSOPTIONcompsoc
\usepackage[nocompress]{cite}
\else
\usepackage{cite}
\fi
\usepackage{caption}

\ifCLASSINFOpdf
\usepackage[pdftex]{graphicx}
\graphicspath{{./fig/}}
\usepackage{epstopdf}
\DeclareGraphicsExtensions{.pdf,.jpeg,.png,.eps}
\else
\usepackage[dvips]{graphicx}
\graphicspath{{./fig/}}
\DeclareGraphicsExtensions{.eps}
\fi

\usepackage[cmex10]{amsmath}
\usepackage{amsmath,amsthm,amssymb,amsfonts,mathrsfs,bm}
\usepackage{algorithm,algorithmicx,algpseudocode,hyperref}
\usepackage{ragged2e}
\usepackage{tikz}
\usepackage{epsfig,bbm}
\usepackage{dsfont}

\usepackage{setspace}
\usepackage[export]{adjustbox}

\ifCLASSOPTIONcompsoc
\usepackage[caption=false,font=footnotesize,labelfont=sf,
textfont=sf]{subfig} 
\else
\usepackage[caption=false,font=footnotesize]{subfig}
\fi

\begin{document}
\title{ Data-adaptive Active Sampling for Efficient\\ Graph-Cognizant Classification }

\author{\textit{Dimitris Berberidis and Georgios B. Giannakis}\\
Dept. of ECE and Digital Tech. Center, University of Minnesota\\
Minneapolis, MN 55455, USA\\
E-mails: \{bermp001,georgios\}@umn.edu
\thanks{Work was supported by NSF 1514056 and 1500713, and NIH 1R01GM104975-01}}

\maketitle

\begin{abstract}The present work deals with active sampling of graph nodes representing training data for binary classification. The graph may be given or constructed using similarity measures among nodal features. Leveraging the graph for classification builds on the premise that labels across neighboring nodes are correlated according to a categorical Markov random field (MRF). This model is further relaxed to a Gaussian (G)MRF with labels taking continuous values - an approximation that not only mitigates the combinatorial complexity of the categorical model, but also offers optimal unbiased soft predictors of the unlabeled nodes. The proposed sampling strategy is based on querying the node whose label disclosure is expected to inflict the largest change on the GMRF, and in this sense it is the most informative on average. Such a strategy subsumes several measures of expected model change, including uncertainty sampling, variance minimization, and sampling based on the $\Sigma-$optimality criterion. A simple yet effective heuristic is also introduced for increasing the exploration capabilities of the sampler, and reducing bias of the resultant classifier, by taking into account the confidence on the model label predictions.  The novel sampling strategies are based on quantities that are readily available without the need for model retraining, rendering them computationally efficient and scalable to large graphs. Numerical tests using synthetic and real data demonstrate that the proposed methods achieve accuracy that is comparable or superior to the state-of-the-art even at reduced runtime.   

\end{abstract}

\section{Introduction}\label{sec:intro}

Active learning has recently gained popularity for various applications ranging from bioinformatics \cite{cancer} to distributed signal classification and estimation \cite{haupt2011distilled}. In contrast to traditional passive supervised and semi-supervised learning methods, where classifiers are trained using given or randomly drawn labeled data, active learning allows for judiciously selecting which data are to be queried and added to the training set. Thus, a
typical active learner begins with a small labeled set, and proceeds to select one or more informative query instances
from a large unlabeled pool. Active learning yields markedly improved classification accuracy over passive or random sampling when the number of training labels is fixed\cite{survey2013,settles2008,jordan96,chaloner1995bayesian}. It can be particularly appealing when unlabeled data (instances) are readily available, but obtaining
training labels is expensive. For instance, a classifier trained to predict the presence of cancer based on certain protein attributes requires labels that involve costly and time-consuming medical examinations; see e.g., \cite{cancer}.   

Even intuitively, one expects active sampling to outperform random sampling when (un)labeled instances are correlated. Such a case emerges with graph-aware classification, where each instance is denoted by a node, while edges capture correlation among connected nodes. Although graphs may arise naturally in certain applications (e.g. social and citation networks), they can in general be constructed from any set of nodal feature vectors using proper similarity measures; see e.g., \cite{gkgg2016icassp,sbg2017tsp}. Knowing the graph topology, \emph{graph-cognizant} classification boils down to propagating the information from labeled nodes to unlabeled ones through edges of neighboring nodes; see e.g., \cite{zhu2003semi}. As a result, classification on graphs is inherently semi-supervised and thus conducive to active learning.

Prior art in graph-based active learning can be divided in two categories. The first includes the \emph{non-adaptive} design-of-experiments-type methods, where sampling strategies are designed \emph{offline} depending only on the graph structure, based on ensemble optimality criteria. The non-adaptive category also includes the variance minimization sampling \cite{VM12}, as well as the error upper bound minimization in \cite{errorbound2012}, and the data non-adaptive $\Sigma$-optimality approach in \cite{sigmaopt}. The second category includes methods
that select samples adaptively and jointly with the classification process, taking into account both graph structure as well as previously obtained labels. Such \emph{data-adaptive} methods give rise to sampling schemes that are not optimal on average, but adapt to a given realization of labels on the graph. Adaptive methods include the Bayesian risk minimization \cite{riskmin2003}, the information gain maximization \cite{IG08}, as well as the manifold preserving method of \cite{manifold14}; see also \cite{nowak16,Ortegacut16,trees13}. Finally, related works deal with selective sampling of nodes that arrive sequentially in a gradually augmented graph  \cite{selective13,streaming16,augment2015}, as well as active sampling to infer the graph structure \cite{edgesampling2012,directed2012}.

In this context, the present work develops data-adaptive pool based active samplers for graph-aware classification. The proposed sampling strategy relies on querying the node that is expected to inflict the largest change on the underlying label correlation model. Albeit in different context, a related criterion was adopted for semantic segmentation of images \cite{output13}, and for regression of Gaussian processes \cite{output14}. The unifying approach here advocates novel metrics of expected model change, but also includes existing methods such as uncertainty sampling, variance minimization and sampling based on the $\Sigma-$optimality criterion. A simple yet effective heuristic is also introduced for improving the exploration capabilities, and for reducing the bias of the resultant classifiers, by taking into account the confidence on the model label predictions. 

The rest of the paper is organized as follows. Section \ref{sec:problem} states the problem, and presents the GMRF model adopted to approximate the marginal distributions of the unknown categorical node labels. Section \ref{sec:main} develops active learning methods based on different measures of change, and establishes links to existing sampling schemes. Section \ref{sec:confidence} points out the issue of sampling bias, and mitigates it by incorporating a confidence metric on the underlying model. Finally, Section \ref{sec:simulations} presents numerical experiments on real and synthetic datasets.

\emph{Notation.} Lower- (upper-) case boldface letters denote column vectors (matrices). Calligraphic symbols are reserved for sets, while $^T$ stands for transposition. Vectors $\mathbf{0}$, $\mathbf{1}$, and $\mathbf{e}_n$ denote the all-zeros, the all-ones, and the $n$-th canonical vector, respectively. Symbol $\mathds{1}_{E}$ denotes the indicator for the event $E$. Notation $\mathcal{N}(\mathbf{m},\mathbf{C})$ stands for the multivariate Gaussian distribution with mean $\mathbf{m}$ and covariance matrix $\mathbf{C}$, while $\mathrm{tr(\mathbf{X})}$, $\lambda_{\min}(\mathbf{X})$, and $\lambda_{\max}(\mathbf{X})$ are reserved for the trace, the minimum and maximum eigenvalues of matrix $\mathbf{X}$, respectively. Symbol $[\mathbf{x}]_i$ denotes the $i-$th entry of vector $\mathbf{x}$.

\section{Modeling and Problem Statement}\label{sec:problem}

Consider a connected undirected graph
$\mathcal{G}=\{\mathcal{V}, \mathcal{E}\}$, where $\mathcal{V}$ is the set of $N$ nodes, and $\mathcal{E}$ contains the edges that are also represented by the $N \times N$ weighted adjacency matrix $\mathbf{W}$ whose $(i,j)-$th entry denotes the weight of the edge that connects nodes $v_i$ and $v_j$. Let us further suppose that a binary label $y_i\in \{-1,1\}$ is associated with each node $v_i$. The weighted binary labeled graph can either be given, or, it can be inferred from a set of $N$ data points $\{ \mathbf{x}_i,y_i \}_{i=1}^N$ such that each node of the graph corresponds to a data point. Matrix $\mathbf{W}$ can be obtained from the feature vectors  $\{\mathbf{x}_i\}_{i=1}^N$ using different similarity measures. For example, one may use the radial basis function 
$
w_{i,j}=\exp\left(-\|\mathbf{x}_i-\mathbf{x}_j\|_2^2/{\sigma^2}\right)
$
that assigns large edge weights to pairs of points that are neighbors in Euclidean space, or the Pearson correlation coefficients 
$
w_{i,j}={\langle \mathbf{x}_i,\mathbf{x}_j \rangle}/\left(\|\mathbf{x}_i\|_2\|\mathbf{x}_j\|_2\right).
$
If $w_{i,j}\neq 0 ~\forall {i,j}$, the resulting graph will be fully connected, but one may obtain a more structured graph by rounding small weights to $0$.

Having embedded the data on a graph, semi-supervised learning amounts to propagating an observed subset of labels to the rest of the network. Thus, upon observing $\{y_i\}_{i\in \mathcal{L}}$ where $\mathcal{L}\subseteq \mathcal{V}$, henceforth collected in the $|\mathcal{L}|\times 1$ vector $\mathbf{y}_{\mathcal{L}}$, the goal is to infer the labels of the unlabeled nodes $\{y_i\}_{i\in \mathcal{U}}$ concatenated in the vector $\mathbf{y}_{\mathcal{U}}$, where $\mathcal{U}:=\mathcal{V} /\mathcal{L}$. Let us consider labels as random variables that follow an unknown joint distribution $(y_1,y_2,\ldots ,y_N)  \sim p(y_1,y_2,\ldots ,y_N)$, or $\mathbf{y}\sim p(\mathbf{y})$ for brevity.

 For the purpose of inferring unobserved from observed labels, it would suffice if the joint posterior distribution
$
p\left( \mathbf{y}_{\mathcal{U}} | \mathbf{y}_{\mathcal{L}} \right)
$
were available; then, $\mathbf{y}_{\mathcal{U}}$ could be obtained as a combination of labels that maximizes $p\left( \mathbf{y}_{\mathcal{U}} | \mathbf{y}_{\mathcal{L}} \right)$. Moreover, obtaining the marginal posterior distributions $p\left( y_i | \mathbf{y}_{\mathcal{L}} \right)$ of each unlabeled node $i$ is often of interest, especially in the present greedy active sampling approach. To this end, it is well documented that MRFs are suitable for modeling probability mass functions over undirected graphs using the generic form, see e.g., \cite{riskmin2003}
\begin{subequations}
\begin{equation}\label{BMRF}
p(\mathbf{y}):=\frac{1}{Z_{\beta}}\exp{(-\frac{\beta}{2} \Phi(\mathbf{y}))}~~~~~~~~~~
\end{equation}
where the ``partition function'' $Z_{\beta}$ ensures that \eqref{BMRF} integrates to $1$, $\beta$ is a scalar that controls the smoothness of $p(\mathbf{y})$, and $\Phi(\mathbf{y})$ is the so termed ``energy'' of a realization $\mathbf{y},$ given by 
\begin{align}\label{energy}
\Phi(\mathbf{y})&:=\sum_{i,j\in \mathcal{V}}w_{i,j}\left(y_i-y_j\right)^2=\mathbf{y}^T\mathbf{L}\mathbf{y}
\end{align}
\end{subequations}
that captures the graph-induced label dependencies through the graph Laplacian matrix $\mathbf{L}:=\mathbf{D}-\mathbf{W}$ with $\mathbf{D}:=\mathrm{diag}(\mathbf{W 1})$. This categorical MRF model in \eqref{BMRF} naturally incorporates the known graph structure (through $\mathbf{L}$) in the label distribution by assuming label configurations where nearby labels (large edge weights) are similar, and have lower energy as well as higher likelihood. Still, finding the joint and marginal posteriors using \eqref{BMRF} and \eqref{energy} incurs \emph{exponential complexity} since $\mathbf{y}_{\mathcal{U}}\in\{-1,1\}^{|\mathcal{U}|}$. To deal with this challenge, less complex continuous-valued models are well motivated for a scalable approximation of the marginal posteriors. This prompts our next step to allow for continuous-valued label configurations $\boldsymbol{\psi}_{\mathcal{U}}\in\mathbb{R}^{|\mathcal{U}|}$ that are modeled by a GMRF.

\subsection{GMRF relaxation}

Consider approximating the binary field $\mathbf{y}\in\{-1,1\}^{|\mathcal{U}|}$ that is distributed according to \eqref{BMRF} with the continuous-valued $\boldsymbol{\psi}\sim \mathcal{N}(\mathbf{0},\mathbf{C})$, where the covariance matrix satisfies $\mathbf{C}^{-1}=\mathbf{L}$.  
 Label propagation under this relaxed GMRF model becomes readily available in closed form. Indeed, $\boldsymbol{\psi}_{\mathcal{U}|\mathcal{L}}$ of unlabeled nodes conditioned on the labeled ones obeys 
\begin{equation}\label{Gauss_field}
\boldsymbol{\psi}_{\mathcal{U}|\mathcal{L}}\sim \mathcal{N}(\boldsymbol{\mu}_{\mathcal{U}|\mathcal{L}},\mathbf{L}_{\mathcal{UU}}^{-1})
\end{equation}
where $\mathbf{L}_{\mathcal{UU}}$ is the part of the graph Laplacian that corresponds to unlabeled nodes in the partitioning 
\begin{equation}
\mathbf{L}=\left[\begin{array}{cc}
\mathbf{L}_{\mathcal{UU}} & \mathbf{L}_{\mathcal{UL}} \\
\mathbf{L}_{\mathcal{LU}} & \mathbf{L}_{\mathcal{LL}} \end{array} \right].
\end{equation}
Given the observed $\boldsymbol{\psi}_{\mathcal{L}}$, the minimum mean-square error (MMSE) estimator of $\boldsymbol{\psi}_{\mathcal{U}}$ is given by the conditional expectation
\begin{align}\nonumber
\boldsymbol{\mu}_{\mathcal{U}|\mathcal{L}}&=\mathbf{C}_{\mathcal{UL}}\mathbf{C}_{\mathcal{LL}}^{-1}\boldsymbol{\psi}_{\mathcal{L}} \\\label{cond_exp}
&=-\mathbf{L}_{\mathcal{UU}}^{-1}\mathbf{L}_{\mathcal{UL}}\boldsymbol{\psi}_{\mathcal{L}}
\end{align}
where the first equality holds because for jointly Gaussian zero-mean vectors the MMSE estimator coincides with the linear (L)MMSE one (see e.g., \cite[p. 382]{kay93book}), while the second equality is derived in Appendix A1.
%
When binary labels $\mathbf{y}_{\mathcal{L}}$ are obtained, they can be treated as measurements of the continuous field ($\boldsymbol{\psi}_{\mathcal{L}}:=\mathbf{y}_{\mathcal{L}}$), and \eqref{cond_exp} reduces to
 \begin{align}\label{cond_exp2}
 \boldsymbol{\mu}_{\mathcal{U}|\mathcal{L}}=-\mathbf{L}_{\mathcal{UU}}^{-1}\mathbf{L}_{\mathcal{UL}}\mathbf{y}_{\mathcal{L}}.
 \end{align}
  Interestingly, the conditional mean of the GMRF in \eqref{cond_exp2} may serve as an approximation of the marginal posteriors of the unknown labels. Specifically, for the $i-$th node, we adopt the approximation
\begin{align}\nonumber
p\left( y_i=1 |\mathbf{y}_{\mathcal{L}} \right)&=\frac{1}{2}\left(\mathbb{E}\left[\big[\mathbf{y}_{\mathcal{U}|\mathcal{L}}]_i\right]+1\right)\\\nonumber
&\approx \frac{1}{2}\left(\mathbb{E}\left[\big[\boldsymbol{\psi}_{\mathcal{U}|\mathcal{L}}\big]_i\right]+1\right)\\\label{posterior_approx}
&:= \frac{1}{2}\left(\big[\boldsymbol{\mu}_{\mathcal{U}|\mathcal{L}}\big]_i+1\right)
\end{align}
 where the first equality follows from the fact that the expectation of a Bernouli random variable equals its probability. Given the approximation of $p\left( y_i |\mathbf{y}_{\mathcal{L}} \right)$ in \eqref{posterior_approx}, and the uninformative prior $p(y_i =1)=0.5~\forall i \in \mathcal{V}$, the maximum a posteriori (MAP) estimate of ${y}_i$, which in the Gaussian case here reduces to the minimum distance decision rule, is given as 
\begin{equation}\label{predict}
\hat{y}_i=
\left\{ \begin{array}{cc}
1~&~\big[\boldsymbol{\mu}_{\mathcal{U}|\mathcal{L}}\big]_i>0\\
-1~&~\mathrm{else}
\end{array} \right. ,~~\forall i \in \mathcal{U}
\end{equation}
thus completing the propagation of the observed $\mathbf{y}_{\mathcal{L}}$ to the unlabeled nodes of the graph.

It is worth stressing at this point, that as the set of labeled samples changes, so does the dimensionality of the conditional mean in \eqref{cond_exp2}, along with the ``auto-" and ``cross-" Laplacian sub-matrices that enable soft label propagation via \eqref{cond_exp2}, and hard label propagation through \eqref{predict}. Two remarks are now in order.

\noindent \textbf{Remark 1}. It is well known that the Laplacian of a graph is not invertible, since $\mathbf{L1}=\mathbf{0}$; see, e.g. \cite{kolaczyk2014statistical}. To deal with this issue, we instead use $\mathbf{L}+\delta\mathbf{I}$, where $\delta\ll 1$ is selected arbitrarily small but large enough to guarantee the numerical stability of e.g.,  $\mathbf{L}_{\mathcal{UU}}$ in \eqref{cond_exp2}. A closer look at the energy function $\Phi(\mathbf{y}):=\sum_{i,j\in \mathcal{V}}w_{i,j}\left(y_i-y_j\right)^2+\delta\sum_{i\in \mathcal{V}}y_i^2$ reveals that this simple modification amounts to adding a ``self-loop" of weight $\delta$ to each node of the graph. {Alternatively, $\delta$ can be viewed as a regularizer that ``pushes'' the entries of the Gaussian field $\boldsymbol{\psi}_{\mathcal{U}}$ closer to $0$, which also causes the (approximated) marginal posteriors $p(y_i|\mathbf{y}_{\mathcal{L}})$ to be closer to $0.5$ (cf. eq. (6)). In that sense, $\delta$ enforces the priors $p(y_i=1)=p(y_i=0)=0.5$.}

\noindent \textbf{Remark 2}. The method introduced here for label propagation (cf. \eqref{cond_exp2}) is related with the one reported in \cite{riskmin2003}. The main differences are: i) we perform soft label propagation by minimizing the mean-square prediction error of unlabeled from labeled samples; and ii) our model approximates $\{-1,1\}$ labels with a \emph{zero-mean} Gaussian field, while the model in \cite{riskmin2003} approximates $\{ 0,1 \}$ labels also with a zero-mean Gaussian field (instead of one centered at $0.5$). Apparently, \cite{riskmin2003} treats the two classes differently since it exhibits a bias towards class $0$; thus, simply denoting class $0$ as class $1$ yields different marginal posteriors and classification results. In contrast, our model is bias-free and treats the two classes equally.

\subsection{Active sampling with GMRFs}

\begin{algorithm}[t!]
	\caption{Active Graph Sampling Algorithm}\label{algorithm}
	\begin{algorithmic}
		\State  \underline{\textbf{Input:}} Adjacency matrix $\mathbf{W}$, $\delta \ll 1$
		\State \underline{\textbf{Initialize:}} $\mathcal{U}^{0}=\mathcal{V}$, $\mathcal{L}^{0}=\emptyset$, $\boldsymbol{\mu}=\mathbf{0}, \mathbf{G}_0=(\mathbf{L}+\delta \mathbf{I})^{-1}$
		\State First query is chosen at random
		\For {$t=1:T$}
		\State Scan $\mathcal{U}^{t-1}$ to find best query node $v_{k_t}$ as in \eqref{greedy}                     
		\State Obtain label $y_{k_t}$ of $v_{k_t}$
		\State Update the GMRF mean as in \eqref{update_mean}
		\State Update $\mathbf{G}_t$ as in \eqref{laplacian_update} 
		\State $\mathcal{U}^{t}=\mathcal{U}^{t-1}/\{{k_t}\}$, $\mathcal{L}^{t}=\mathcal{L}^{t-1}\cup\{{k_t}\}$
		\EndFor
		\State Predict remaining unlabeled nodes as in  \eqref{predict}
	\end{algorithmic}
\end{algorithm}

In passive learning, $\mathcal{L}$ is either chosen at random, or, it is determined a priori. In our pool based active learning setup, the learner can examine a set of instances (nodes in the context of graph-cognizant classification), and can choose which instances to label. Given its cardinality $|\mathcal{L}|$, one way to approximate the exponentially complex task of selecting $\mathcal{L}$ is to \emph{greedily} sample one node per iteration $t$ with index  
\begin{equation}\label{greedy}
{k_t}=\arg\max_{i\in \mathcal{U}^{t-1}}{U}(v_i,\mathcal{L}^{t-1})
\end{equation}
where ${U}(v,\mathcal{L}^{t-1})$ is a utility function that evaluates how informative node $v$ is while taking into account information already available in $\mathcal{L}^{t-1}$. Upon disclosing label $y_{k_t}$, it can be shown that instead of re-solving \eqref{cond_exp2}, the GMRF mean can be updated recursively using the ``dongle node" trick in \cite{riskmin2003} as 
\begin{equation}\label{update_mean}
\boldsymbol{\mu}_{\mathcal{U}^{t-1}|\mathcal{L}^{t-1}}^{+y_{k_t}}=\boldsymbol{\mu}_{\mathcal{U}^{t-1}|\mathcal{L}^{t-1}} + \frac{1}{g_{{k_t}{k_t}}}(y_{k_t}-[\boldsymbol{\mu}_{\mathcal{U}^{t-1}|\mathcal{L}^{t-1}}]_{k_t})\mathbf{g}_{k_t}
\end{equation}
where $\boldsymbol{\mu}_{\mathcal{U}^{t-1}|\mathcal{L}^{t-1}}^{+y_{k_t}}$ is the conditional mean of the unlabeled nodes when node $v_{k_t}$ is assigned label $y_{k_t}$ (thus ``gravitating" the GMRF mean $[\boldsymbol{\mu}_{\mathcal{U}^{t-1}|\mathcal{L}^{t-1}}]_{k_t}$ toward its replacement $y_{k_t}$); vector $\mathbf{g}_{k_t}:=[\mathbf{L}_{\mathcal{U}^{t-1}\mathcal{U}^{t-1}}^{-1}]_{:{k_t}}$ and scalar $g_{{k_t}{k_t}}:=[\mathbf{L}_{\mathcal{U}^{t-1}\mathcal{U}^{t-1}}^{-1}]_{{k_t}{k_t}}$ are the ${k_t}-$th column and diagonal entry of the Laplacian inverse, respectively.  {Subsequently, the new conditional mean vector $\mu_{\mathcal{U}^{t}|\mathcal{L}^{t}}$ defined over $\mathcal{U}^{t}$ is given by removing the $i-$th entry of $\mu_{\mathcal{U}^{t-1}|\mathcal{L}^{t-1}}^{+y_i}$. }
		 Using Shur's lemma it can be shown that the inverse Laplacian $\mathbf{G}_t^{-{k_t}}$ when the ${k_t}-$th node is removed from the unlabeled sub-graph can be efficiently updated from $\mathbf{G}_t:=\mathbf{L}_{\mathcal{U}^{t}\mathcal{U}^{t}}^{-1}$ as \cite{sigmaopt}
\begin{equation}\label{laplacian_update}
\left[\begin{array}{cc}
\mathbf{G}_t^{-{k_t}} & \mathbf{0} \\
\mathbf{0}^T & 0 \end{array} \right] = \mathbf{G}_t - \frac{1}{g_{{k_t}{k_t}}} \mathbf{g}_{k_t} \mathbf{g}_{k_t}^T 
\end{equation}
which requires only $\mathcal{O}(|\mathcal{U}|^2)$ computations instead of $\mathcal{O}(|\mathcal{U}|^3)$ for matrix inversion. Alternatively, one may obtain $\mathbf{G}_t^{-{k_t}}$ by applying the matrix inversion lemma employed by the RLS-like solver in \cite{riskmin2003}. The resultant greedy active sampling scheme for graphs is summarized in Algorithm \ref{algorithm}.

\noindent  \textbf{Remark 3}. Existing data-adaptive sampling schemes, e.g., \cite{riskmin2003}, \cite{nowak16}, \cite{IG08}, often require \emph{model-retraining} by examining candidate labels per unlabeled node (cf. \eqref{greedy}). Thus, even when retraining is efficient, it still needs to be performed $|\mathcal{U}||\mathcal{C}|$ times per iteration of Algorithm 1, which in practice significantly increases runtime, especially for larger graphs. 

 In summary, different sampling strategies emerge by selecting distinct utilities ${U}(v,\mathcal{L}^{t-1})$ in \eqref{greedy}. In this context, the goal of the present work is to develop novel active learning schemes based on a unifying sampling approach that subsumes existing alternatives, and can mitigate their sampling bias. A further desirable attribute of the sought approach is to bypass the need for GMRF retraining.

\section{Expected model change}\label{sec:main}

Judiciously selecting the utility function is of paramount importance in designing an efficient active sampling algorithm. In the present work, we introduce and investigate the relative merits of different choices under the prism of expected change (EC) measures that we advocate as information-revealing utility functions. From a high-level vantage point, the idea is to identify and sample nodes of the graph that are expected to have the greatest impact on the available GMRF model of the unknown labels. Thus, contrary to the expected error reduction and entropy minimization approaches that actively sample with the goal of increasing the ``confidence'' on the model, our focus is on effecting maximum perturbation of the model with each node sampled. The intuition behind our approach is that by sampling nodes with large impact, one may take faster steps towards an increasingly accurate model.      

\subsection{EC of model predictions}

An intuitive measure of expected model change for a given node $v_i$ is the expected number of unlabeled nodes whose label prediction will change after sampling the label of $v_i$. To start, consider per node $i$ the measure
\begin{equation}\label{flips}
F(y_i,\boldsymbol{\mu}_{\mathcal{U}|\mathcal{L}}):=\sum_{j\in \mathcal{U}-\{i\}}\mathds{1}_{\{\hat{y}_j^{+y_i}\neq \hat{y}_j\}}
\end{equation}
denote the number of such ``flips'' in the predicted labels of \eqref{predict}. For notational brevity, we henceforth let  $\mu_i=[\boldsymbol{\mu}_{\mathcal{U}|\mathcal{L}}]_i$. The corresponding utility function is 
\begin{align}\nonumber
U_{FL}(v_i,\mathcal{L})&=\mathbb{E}_{y_i|\mathbf{y}_{\mathcal{L}}}\left[F(y_i,\boldsymbol{\mu}_{\mathcal{U}|\mathcal{L}})\right]\\\nonumber
&=p(y_i=1|\mathbf{y}_{\mathcal{L}})F(y_i=1,\boldsymbol{\mu}_{\mathcal{U}|\mathcal{L}})\\\nonumber
&+p(y_i=-1|\mathbf{y}_{\mathcal{L}})F(y_i=-1,\boldsymbol{\mu}_{\mathcal{U}|\mathcal{L}})\\\nonumber
&\approx \frac{1}{2}(\mu_i+1)F(y_i=1,\boldsymbol{\mu}_{\mathcal{U}|\mathcal{L}})\\\label{Expected_flips}
&+\left(1-\frac{1}{2}(\mu_i+1)\right)F(y_i=-1,\boldsymbol{\mu}_{\mathcal{U}|\mathcal{L}})
\end{align}
where the approximation is because \eqref{posterior_approx} was used in place of $p(y_i=1|\mathbf{y}_{\mathcal{L}})$. Note that model retraining using \eqref{update_mean} is required to be performed twice (in general, as many as the number of classes) for each node in $\mathcal{U}$ in order to obtain the labels $\{\hat{y}_j\}^{+y_i}$ in \eqref{flips}.

\subsection{EC using KL divergence}
The utility function in \eqref{Expected_flips} depends on the hard label decisions of \eqref{predict}, but does not account for perturbations that do not lead to changes in label predictions. To obtain utility functions that are more sensitive to the soft GMRF model change, it is prudent to measure how much the continuous distribution of the unknown labels changes after sampling. Towards this goal, we consider first the KL divergence between two pdfs $p(\mathbf{x})$ and $q(\mathbf{x})$, which is defined as
	\begin{align}\nonumber
	\mathcal{D}_{KL}(p||q):=\int p(\mathbf{x}) \ln \frac{p(\mathbf{x})}{q(\mathbf{x})} d \mathbf{x} =\mathbb{E}_{p}\left[\ln \frac{p(\mathbf{x})}{q(\mathbf{x})}\right].
	\end{align}
For the special case where $p(\mathbf{x})$ and $q(\mathbf{x})$ are Gaussian with identical covariance matrix $\mathbf{C}$ and corresponding means $\mathbf{m}_p$ and $\mathbf{m}_q$, their KL divergence is expressible in closed form as 
\begin{equation}\label{normalKL}
\mathcal{D}_{KL}(p||q):=\frac{1}{2}(\mathbf{m}_p-\mathbf{m}_q)^T\mathbf{C}^{-1}(\mathbf{m}_p-\mathbf{m}_q)
\end{equation}

Upon recalling that $\boldsymbol{\psi}_{\mathcal{U}}$ defined over the unlabeled nodes is Gaussian [cf. \eqref{Gauss_field}], and since the Gaussian field obtained after node $v_i$ is designated label $y_i$ is also Gaussian, we have
\begin{equation}
\boldsymbol{\psi}_{\mathcal{U}}^{+ y_i }\sim \mathcal{N}(\boldsymbol{\mu}_{\mathcal{U}|\mathcal{L}}^{+y_i},\mathbf{L}^{-1}_{\mathcal{UU}}).
\end{equation}
It thus follows that the KL divergence induced on the GMRF after sampling $y_i$ is (cf. \eqref{normalKL})
\begin{align}\nonumber
\mathcal{D}_{KL}(\boldsymbol{\psi}_{\mathcal{U}}^{+ y_i }||\boldsymbol{\psi}_{\mathcal{U}})&=\frac{1}{2}\left[ (\boldsymbol{\mu}_{\mathcal{U}|\mathcal{L}}^{+y_i}-\boldsymbol{\mu}_{\mathcal{U}|\mathcal{L}})^T\mathbf{L}_{\mathcal{UU}}(\boldsymbol{\mu}_{\mathcal{U}|\mathcal{L}}^{+y_i}-\boldsymbol{\mu}_{\mathcal{U}|\mathcal{L}}) \right]\\\label{Gauss_KL}
&=\frac{1}{2g_{ii}^2}(y_i - \mu_i)^2{\mathbf{g}_i^T\mathbf{L}_{\mathcal{UU}}\mathbf{g}_i}=\frac{1}{2{g_{ii}}}{(y_i - \mu_i)^2}
\end{align}
where the second equality relied on \eqref{update_mean}, and the last equality used the definition of $g_{ii}$. The divergence in \eqref{Gauss_KL} can be also interpreted as the normalized innovation of observation $y_i$. Averaging \eqref{Gauss_KL} over the candidate values of $y_i$ yields the expected KL divergence of the GMRF utility as
\begin{align}\nonumber
U_{KLG}(v_i,\mathcal{L})&=\mathbb{E}_{y_i|\mathbf{y}_{\mathcal{L}}}\left[\mathcal{D}_{KL}(\boldsymbol{\psi}_{\mathcal{U}}^{+ y_i }||\boldsymbol{\psi}_{\mathcal{U}})\right]\\\nonumber
&=p(y_i=1|\mathbf{y}_{\mathcal{L}})\mathcal{D}_{KL}(\boldsymbol{\psi}_{\mathcal{U}}^{+ y_i=1 }||\boldsymbol{\psi}_{\mathcal{U}})\\\nonumber
&+p(y_i=-1|\mathbf{y}_{\mathcal{L}})\mathcal{D}_{KL}(\boldsymbol{\psi}_{\mathcal{U}}^{+ y_i=-1 }||\boldsymbol{\psi}_{\mathcal{U}})\\\nonumber
&\approx \frac{1}{2}\bigg[\frac{1}{2{g_{ii}}}(\mu_i+1){(1 - \mu_i)^2}\\\nonumber
&+\left(1-\frac{1}{2}(\mu_i+1)\right)\frac{1}{g_{ii}}(-1-\mu_i)^2\bigg]\\\label{KL_1}
&=\frac{1}{2g_{ii}}(1-\mu_i^2).
\end{align}
Interestingly, the utility in \eqref{KL_1} leads to a form of uncertainty sampling, since $(1-\mu_i^2)$ is a measure of uncertainty of the model prediction for node $v_i$, further normalized by $g_{ii}$, which is the variance of the Gaussian field (cf. \cite{VM12}).
Note also that the expected KL divergence in \eqref{KL_1} also relates to the information gain between $\{\psi_j\}_{j \in \mathcal{U}/\{ i \}}$ and $y_i$.

Albeit easy to compute since model retraining is not required, ${U_{KLG}}$ quantifies the impact of disclosing $y_i$ on the GMRF, but not the labels $\{y_j\}_{j \in \mathcal{U}/\{ i \}}$ themselves. To account for the labels themselves, an alternative KL-based utility function could treat $\{y_j\}_{j \in \mathcal{U}-\{ i \}}$  as Bernouli variables [c.f. \eqref{posterior_approx}]; that is  
{
\begin{align}
y_j\sim \mathrm{Ber}((\mu_j+1)/2).
\end{align}
}
In that case, one would ideally be interested in obtaining the expected KL divergence between the true posteriors, that is
\begin{align}\label{ideal}
\mathbb{E}_{y_i|\mathbf{y}_{\mathcal{L}}}\left[\mathcal{D}_{KL}\left(p({\mathbf{y}}_{\mathcal{U}}|\mathbf{y}_{\mathcal{L}},y_i)||p({\mathbf{y}}_{\mathcal{U}}|\mathbf{y}_{\mathcal{L}})\right)\right].
\end{align}
Nevertheless, the joint pdfs of the labels are not available by the GMRF model; in fact, any attempt at evaluating the joint posteriors incurs exponential complexity as mentioned in Section \ref{sec:problem}. {One way to bypass this limitation is by approximating the joint posterior $p({\mathbf{y}}_{\mathcal{U}}|\mathbf{y}_{\mathcal{L}})$ with the product of marginal posteriors $\prod_{j \in \mathcal{U}}p({\mathbf{y}}_{j}|\mathbf{y}_{\mathcal{L}})$, since the later are readily given by the GMRF. Using this independence assumption causes the joint KL divergence in \eqref{ideal} to break down to the sum of marginal per-unlabeled-node KL divergences. The resulting utility function can be expressed as}
\begin{align}\label{KL_2}
U_{KL}(v_i,\mathcal{L})= \sum_{j\in \mathcal{U}/\{i\}}I(y_j,y_i)
\end{align}
where 
\begin{align}\nonumber
I(y_j,y_i)&=\mathbb{E}_{y_i|\mathbf{y}_{\mathcal{L}}}\left[\mathcal{D}_{KL}\left(p(y_j|\mathbf{y}_{\mathcal{L}},y_i)||p(y_j|\mathbf{y}_{\mathcal{L}})\right)\right]\\\nonumber
&\approx \frac{1}{2}(\mu_i+1)\mathcal{D}_{KL}(y_j^{+ y_i=1 }||y_j)\\\label{KL_3}
&+\left(1-\frac{1}{2}(\mu_i+1)\right)\mathcal{D}_{KL}(y_j^{+ y_i=-1 }||y_j)
\end{align}
since for univariate distributions the expected KL divergence between the prior and posterior is equivalent to the mutual information between the observed random variable and its unknown label. Note also that the KL divergence between univariate distributions is simply
\begin{align}\label{KL_4}
\mathcal{D}_{KL}(y_j^{+ y_i }||y_j)=H(y_j^{+ y_i },y_j)-H(y_j^{+ y_i })
\end{align}
where $H(y_j^{+ y_i },y_j)$ denotes the cross-entropy, which for Bernouli variables is
\begin{align}\nonumber
H(y_j^{+ y_i },y_j)&= -\frac{1}{2}(\mu_j^{+y_i}+1)\log \frac{1}{2}(\mu_j+1) \\\label{KL_5}
&- \left[1-\frac{1}{2}(\mu_j^{+y_i}+1)\right]\log \left[1-\frac{1}{2}(\mu_j+1)\right]. 
\end{align}
Combining \eqref{KL_2}-\eqref{KL_5} yields $U_{KL}$. Intuitively, this utility promotes the sampling of nodes that are expected to induce large change on the model (cross-entropy between old and new distributions), while at the same time increasing the ``confidence'' on the model (negative entropy of updated distribution). { Furthermore, the mutual-information-based expressions (19) and (20) establish a connection to the information-based metrics in \cite{IG08} and \cite{krause2008near}, giving an expected-model-change interpretation of the entropy reduction method.}

\subsection{EC without model retraining}

{In this section, we introduce two measures of model change that do not require model retraining (cf. Remark 3), and hence are attractive in their simplicity. Specifically,  retraining (i.e., computing $\boldsymbol{\mu}_{\mathcal{U}^{t-1}|\mathcal{L}^{t-1}}^{+y_{i}},~ \forall i \in \mathcal{U}$ and $\forall y_i\in\mathcal{Y}$) is not required if per-node utility ${U}(v,\mathcal{L}^{t-1})$ can be given in \emph{closed-form} as a function of $\mathbf{G}_{t-1}$ and $\boldsymbol{\mu}_{\mathcal{U}^{t-1}|\mathcal{L}^{t-1}}$. Two such measures are explored here: one based on the sum of total variations that a new sample inflicts on the (approximate) marginal distributions of the unknown labels, and one based on the mean-square deviation that a new sample is expected to inflict on the GMRF. }

The \emph{total variation} between two probability distributions $p(x)$ and $q(x)$ over a finite alphabet $\mathcal{X}$ is  
\begin{align}\nonumber
\delta(p,q):=\frac{1}{2}\sum_{x\in\mathcal{X}}|p(x)-q(x)|.
\end{align}
Using the approximation in \eqref{posterior_approx}, the total variation between the distribution of an unknown label $y_j$ and the same label $y_j^{+y_i}$ after $y_i$ becomes available is 
\begin{align}\nonumber
\delta(y_j^{+y_i},y_j)&=\frac{1}{2}\left(|\mu_j^{+y_i}-\mu_j|+|1-\mu_j^{+y_i}-(1-\mu_j)|\right)\\\label{TV_1}
&=|\mu_j^{+y_i}-\mu_j|.
\end{align}
Consequently, the sum of total variations over all the unlabeled nodes $\{v_j\}_{j\in \mathcal{U}/[\{i\}]}$ is 
\begin{align}\nonumber
\Delta(\mathbf{y}_{\mathcal{U}}^{+ y_i },\mathbf{y}_{\mathcal{U}}) &= \sum_{j \in \mathcal{U}}\delta(y_j^{+y_i},y_j)=\|\boldsymbol{\mu}_{\mathcal{U}|\mathcal{L}}^{+y_i} - \boldsymbol{\mu}_{\mathcal{U}|\mathcal{L}}\|_1\\\nonumber
&=\frac{1}{g_{ii}}|y_i-\mu_i|\|\mathbf{g}_i\|_1
\end{align}
where the second equality follows by concatenating all total variations (cf. \eqref{TV_1}) in vector form, and the last one follows by the GMRF update rule in \eqref{update_mean}. 
Finally, the expected sum of total variations utility score-function is defined as
\begin{align*}
U_{TV}(v_i,\mathcal{L})&:=\mathbb{E}_{y_i|\mathbf{y}_{\mathcal{L}}}\left[\Delta(\mathbf{y}_{\mathcal{U}}^{+ y_i },\mathbf{y}_{\mathcal{U}})\right]\\
&=\mathbb{E}_{y_i|\mathbf{y}_{\mathcal{L}}}\left[|y_i-\mu_i|\right]\frac{1}{g_{ii}}\|\mathbf{g}_i\|_1
\end{align*}
and since
\begin{align*}
\mathbb{E}_{y_i|\mathbf{y}_{\mathcal{L}}}\left[|y_i-\mu_i|\right]
&=p(y_i=1|\mathbf{y}_{\mathcal{L}})|1-\mu_i|\\
&+p(y_i=-1|\mathbf{y}_{\mathcal{L}})|-1-\mu_i|\\
&\approx 2(1-\mu_i^2)
\end{align*}
it follows that the utility function based on total variation can be expressed as
\begin{align}\label{TV}
U_{TV}(v_i,\mathcal{L})=\frac{2}{g_{ii}}(1-\mu_i^2)\|\mathbf{g}_{i}\|_1.
\end{align}

The second measure is based on the \emph{mean-square deviation} (MSD) between two rv's $X_1$ and $X_2$ 
\begin{align*}
\mathrm{MSD}(X_1,X_2)&:=\int \left(X_1-X_2\right)^2 f(X_1,X_2) dX_1dX_2\\
&=\mathbb{E}\left[\left(X_1-X_2\right)^2\right].
\end{align*}
Our next proposed utility score is the expected MSD between the Gaussian fields $\boldsymbol{\psi}_{\mathcal{U}}$ and $\boldsymbol{\psi}_{\mathcal{U}}^{+y_i}$ before and after obtaining $y_i$;  that is,  
\begin{align}\nonumber
U_{\mathit{MSD}}(v_i,\mathcal{L})&=\mathbb{E}_{y_i|\mathbf{y}_{\mathcal{L}}}\left[\mathrm{MSD}(\boldsymbol{\psi}_{\mathcal{U}}^{+ y_i },\boldsymbol{\psi}_{\mathcal{U}})\right]\\\label{MSD_1}
&\approx \frac{1}{2}(\mu_i+1)\mathrm{MSD}(\boldsymbol{\psi}_{\mathcal{U}}^{+ y_i=1 },\boldsymbol{\psi}_{\mathcal{U}})\\\nonumber
& +\left(1-\frac{1}{2}(\mu_i+1)\right)\mathrm{MSD}(\boldsymbol{\psi}_{\mathcal{U}}^{+ y_i=-1 },\boldsymbol{\psi}_{\mathcal{U}})
\end{align}
where 
\begin{align}\nonumber
\mathrm{MSD}(\boldsymbol{\psi}_{\mathcal{U}}^{+ y_i },\boldsymbol{\psi}_{\mathcal{U}})&:=\mathbb{E}\left[\|\boldsymbol{\psi}_{\mathcal{U}}^{+ y_i }-\boldsymbol{\psi}_{\mathcal{U}}\|^2\right]\\\nonumber
&= 2\mathrm{tr}(\mathbf{L}_{\mathcal{UU}}^{-1})+ \|\boldsymbol{\mu}_{\mathcal{U}|\mathcal{L}}^{+y_i}-\boldsymbol{\mu}_{\mathcal{U}|\mathcal{L}}\|_2^2\\\label{MSD_2}
&\propto \frac{1}{g_{ii}^2}(y_i-\mu_i)^2\|\mathbf{g}_i\|_2^2.
\end{align}
The second equality in \eqref{MSD_2} is derived in Appendix A2 under the assumption that  $\boldsymbol{\psi}_{\mathcal{U}}$ and $\boldsymbol{\psi}_{\mathcal{U}}^{+y_i}$ are independent random vectors. Furthermore, the term $2\mathrm{tr}(\mathbf{L}_{\mathcal{UU}}^{-1})$ is ignored since it does not depend on $y_i$, and the final expression of \eqref{MSD_2} is obtained using \eqref{update_mean}. Finally, substituting \eqref{MSD_2} into \eqref{MSD_1} yields the following closed-form expression of the MSD-based utility score function
\begin{align}\label{MSD}
U_{\mathit{MSD}}(v_i,\mathcal{L})\propto (1-\mu_i^2) \frac{\|\mathbf{g}_i\|_2^2}{g_{ii}^2}.
\end{align}

Note that $U_{TV}$ and $U_{\mathit{MSD}}$ are proportional to the expected KL divergence of the Gaussian field $U_{KLG}$ in the previous section since
\begin{align}
U_{TV}(v_i,\mathcal{L})\propto U_{KLG}(v_i,\mathcal{L})\|\mathbf{g}_{i}\|_1
\end{align}
and
\begin{align}
U_{\mathit{MSD}}(v_i,\mathcal{L})\propto U_{KLG}(v_i,\mathcal{L})\|\mathbf{g}_{i}\|_2
\end{align}
 with the norms $\|\mathbf{g}_{i}\|_1$ and $\|\mathbf{g}_{i}\|_2$ quantifying the average influence of the $i-$th node over the rest of the unlabeled nodes.  

It is worth mentioning that our TV- and MSD-based methods subsume the $\Sigma-$optimality-based active learning \cite{sigmaopt} and the variance minimization \cite{VM12} correspondingly. This becomes apparent upon recalling that $\Sigma-$optimality and variance-minimization utility score functions are respectively given by
\begin{equation*}
 U_{\Sigma-opt}(v_i)=\frac{\|\mathbf{g}_i\|_1^2}{g_{ii}}
\end{equation*}
and
\begin{equation*}
U_{VM}(v_i):=\frac{\|\mathbf{g}_i\|_2^2}{g_{ii}}.
\end{equation*}
Then, further inspection reveals that the metrics are related by 
 \begin{equation*}
U_{TV}(v_i)\propto\frac{1}{g_{ii}}(1-\mu_i^2)U_{\Sigma-opt}(v_i)
 \end{equation*}
 and correspondingly
 \begin{equation*}
U_{\mathit{MSD}}(v_i)\propto\frac{1}{g_{ii}}(1-\mu_i^2)U_{VM}(v_i).
 \end{equation*}
  In fact, $U_{TV}$ and $U_{\mathit{MSD}}$ may be interpreted as \emph{data-driven} versions of $U_{\Sigma-opt}$ and $U_{VM}$ that are enhanced with the uncertainty term $(1-\mu_i^2)$. On the one hand, $U_{\Sigma-opt}$ and $U_{VM}$ are design-of-experiments-type methods that rely on ensemble criteria and offer \emph{offline} sampling schemes more suitable for applications where the set $\mathcal{L}$ of nodes may \emph{only} be labeled as a \emph{batch}. On the other hand, $U_{TV}$ and $U_{\mathit{MSD}}$ are data-adaptive sampling schemes that adjust to the \emph{specific realization} of labels, and are expected to outperform their batch counterparts in general. { This connection is established due to $U_{VM}(v_i)$ and $U_{\Sigma-opt}(v_i)$ being $l_2$ and $l_1$ \emph{ensemble} loss metrics on the GMRF (see equations 2.3 and 2.5 in [12]); similarly, MSD (mean square deviation) and TV (total variation) are also $l_2$ (on the GMRF distribution) and $l_1$ (on the binary labels pmf) metrics of \emph{change}.}

 \noindent  \textbf{Remark 4}. While the proposed methods were developed for binary classification, they can easily be modified to cope with multiple classes using the one-vs-the-rest trick. Specifically, for any set $\mathcal{C}$ of possible classes, it suffices to solve $|\mathcal{C}|$ binary problems, each one focused on detecting the presence or absence of a class. Consequently, the maximum among the GMRF means $\mu_i^{(c)}~ \forall c\in \mathcal{C}$ reveals which class is the most likely for the $i-$th node. In addition, the marginal posteriors are readily given by normalizing $\mu_i^{(c)}$'s, that is $$p(y_i=c)=\bar{\mu}_i^{(c)}=\frac{\mu_i^{(c)}}{\sum_{c\in\mathcal{C}}\mu_i^{(c)}}.$$ Using this approximation, the TV-based scheme can be generalized to 
  \begin{align}\label{TV_multi}
  U_{TV}(v_i,\mathcal{L})\propto{\sum_{c\in\mathcal{C}}\left[1-(\bar{\mu}_i^{(c)})^2\right]}\frac{\|\mathbf{g}_{i}\|_1}{g_{ii}}
  \end{align}
 and similarly for the MSD-based scheme.

{
	           \begin{table*}[t]\label{tab:complexity}
	           	\caption{ Computational and memory complexity of various methods }
	           	\begin{center}
	           		\begin{tabular} {|c|c|c|c|c|}
	           			\hline
	           			& Offline & Sampling  & Update & Memory \\
	           			\hline 
	           			Random   & $\mathcal{O}( |\mathcal{E}| |\mathcal{C}| )$   &  $\ast$ & $\ast$ & $\mathcal{O}(|\mathcal{E}| + N|\mathcal{C}| )$ \\
	           			\hline 
	           			VM \cite{VM12}, $\Sigma$-opt \cite{sigmaopt}  & $\mathcal{O}( |\mathcal{L}|N^2 )$   &  $\ast$ & $\ast$ & $\mathcal{O}(N^2)$ \\
	           			\hline
	           			Uncertainty (min. margin) &  $\ast$ & $\mathcal{O}(\log |\mathcal{C}|N)$ & $\mathcal{O}(|\mathcal{E}||\mathcal{C}|)$ & $\mathcal{O}(|\mathcal{E}| + N|\mathcal{C}|)$ \\
	           			\hline 
	           			EER \cite{zhu2003semi}, TSA \cite{nowak16} & $\mathcal{O}(|\mathcal{E}|N)$  & $\mathcal{O}(|\mathcal{C}|^2N^2)$ & $\mathcal{O}(N^2)$ & $\mathcal{O}(N^2)$ \\
	           			\hline
	           			FL  & $\mathcal{O}(|\mathcal{E}|N)$ & $\mathcal{O}(|\mathcal{C}|N^2)$ & $\mathcal{O}(N^2)$ & $\mathcal{O}(N^2)$\\
	           			\hline
	           			TV, MSD & $\mathcal{O}(|\mathcal{E}|N)$ & $\mathcal{O}(|\mathcal{C}|N)$ & $\mathcal{O}(N^2)$ & $\mathcal{O}(N^2)$\\
	           			\hline
	           		\end{tabular}
	           	\end{center}
	           	\vspace{-0.5cm}
	           \end{table*}
\subsection{Computational Complexity analysis}
The present section analyzes the computational complexity of implementing the proposed adaptive sampling methods, as well as that of other common adaptive and non-adaptive active learning approaches on graphs. Complexity here refers to float-point multiplications and is given in $\mathcal{O}(\cdot)$ notation as function of the number of nodes $N$, number of edges $|\mathcal{E}|$ and number of classes $|\mathcal{C}|$. Three types of computational tasks are considered separately: computations that can be performed offline (e.g., initialization), computations required to update model after a new node is observed (only for adaptive methods), and the complexity of selecting a new node to sample (cf. eq. \eqref{greedy}).  
}

{
 Let as begin with the ``plain-vanilla'' label propagation scenario where nodes are randomly (or passively) sampled. In that case, the online framework described in Algorithm 1 and Section II.B is not necessary and the nodes can be classified offline after collecting $|\mathcal{L}|$ samples and obtaining \eqref{cond_exp2} for each class in $\mathcal{C}$. Exploiting the sparsity of the $\mathbf{L}$, \eqref{cond_exp2} can be approximated via a Power-like iteration (see, e.g., \cite{labelprop2012}) with $\mathcal{O}( |\mathcal{E}| |\mathcal{C}| )$ complexity. Similarly to passive sampling, non-adaptive approaches such as the variance-minimization (VM) in \cite{VM12} and $\Sigma$-opt design in \cite{sigmaopt} can also be implemented offline. However, unlike passive sampling, the non-adaptive sampling methods require computation of $\mathbf{G}_0=(\mathbf{L}+\delta \mathbf{I})^{-1}$, which can be approximated with $\mathcal{O}(|\mathcal{E}|N)$ multiplications via the Jacobi method. The offline complexity of VM and $\Sigma$-opt is dominated by the complexity required to design the label set $\mathcal{L}$ which is equivalent to  $|\mathcal{L}|$ iterations of Algorithm 1 using $U_{VM}(v_i)$ and $U_{\Sigma-opt}(v_i)$ correspondingly. Thus, the total offline complexity of VM and $\Sigma$-opt  is $\mathcal{O}( |\mathcal{L}|N^2 )$, while $\mathcal{O}(N^2 )$ memory is required to store and process  $\mathbf{G}_0$.
}

{
In the context of adaptive methods, computational efficiency largely depends on whether matrix $\mathbf{G}$ is used for sampling and updating. Simple methods such as uncertainty sampling based on minimum margin do not require $\mathbf{G}$ and have soft labels updated after each new sample using iterative label-propagation (see, e.g., \cite{IG08}) with $\mathcal{O}(|\mathcal{E}||\mathcal{C}|)$ complexity. Uncertainty-sampling-based criteria are also typically very lightweight requiring for instance sorting class-wise the soft labels of each node ($\mathcal{O}(\log |\mathcal{C}|N)$ per sample). While uncertainty-based methods are faster and more scalable, their accuracy is typically significantly lower than that of more sophisticated methods that use $\mathbf{G}$. Methods that use $\mathbf{G}$ such as the proposed EC algorithms in Section III, the expected-error minimization (EER) in \cite{riskmin2003}, and the two-step approximation (TSA) algorithm in \cite{nowak16} all require $\mathcal{O}(N^2)$ to perform the update in \eqref{laplacian_update}. However, TSA and EER use retraining (cf. Remark 3) which entails $\mathcal{O}(|\mathcal{C}|^2N^2)$ cost  in order to perform one sample selection; in contrast, the proposed MSD and TV methods (cf. \eqref{TV}, \eqref{MSD}) only require $\mathcal{O}(|\mathcal{C}|N)$ for sampling. Note that, the performance  gap between EER and TSA on the one hand and TV and MSD on the other becomes larger as the number of classes $|\mathcal{C}|$ increases.  
}

{ The complexity analysis is summarized in Table I and indicates that the proposed \emph{retraining-free} adaptive methods have lower overall complexity than EER and TSA.
 Nevertheless, before proceeding to numerical tests, an important modification is proposed in the ensuing section in order to deal with the challenge of \emph{bias} that is inherent to all data-adaptive sampling schemes.
}
 
\section{Incorporating model confidence}\label{sec:confidence}

It has been observed that all data-adaptive active learning algorithms are more or less prone to yielding ``biased'' sampling schemes, simply because future sample locations are determined by past samples \cite{settles2012active}. It is also known that uncertainty sampling can be particularly sensitive to bias due to the fact that it is more ``myopic," in the sense that it does not take into account the effect of a potential sample on the generalization capabilities of the classifier. Since the TV- and MSD-based utility score functions in \eqref{TV} and \eqref{MSD_2} are influenced by the uncertainty factor $(1-\mu_i^2)$, it is important to mitigate sampling bias before testing the performance of the proposed approaches. 

Let us begin by observing that most active learning methods, including those based on EC we introduced here, are based on utility score functions that take the general form
\begin{align}\label{averaging}
U(v_i,\mathcal{L})=\mathbb{E}_{y_i|\mathbf{y}_{\mathcal{L}}}\left[C(y_i,\mathcal{L})\right]
\end{align}
where $C(y_i,\mathcal{L})$ is any metric that evaluates the effect of node $v_i$ on the model, \emph{given that} its label is $y_i$.
Using the existing probability model to predict how the model itself will change, induces ``bias'' especially in the early stages of the sampling process when the inferred model is most likely far from the true distribution.

 One possible means of reducing bias is by complementing greedy active learning strategies with random sampling. Thats is, instead of selecting the index ${k_t}$ of the node to be sampled at the $t-$th iteration according to \eqref{greedy}, one can opt for a two-branch hybrid rule
\begin{align}\label{mixing}
{k_t}=\bigg\{
\begin{array}{cc}
 \arg\max_{i\in \mathcal{U}^{t-1}}{U}(v_i,\mathcal{L}^{t-1}), & \mathrm{~~~w.p.~~~} (1- \pi^t) \\
\mathrm{Unif}\{1,\ldots, |\mathcal{L}^{t-1}|\}, & \mathrm{w.p.~~~~~~~}\pi^t
\end{array}.
\end{align}
where $\pi^t$ is the probability that at iteration $t$ the sampling strategy switches to uniform random sampling over the unlabeled nodes. Naturally, one should generally select a sequence $\{\pi^t\}$ such that $\pi^t\rightarrow 0$ as $t$ increases the model becomes more accurate. Upon testing the simple heuristic in \eqref{mixing} we observed that it can significantly improve the performance of the more ``myopic'' active sampling strategies. Specifically, uncertainty sampling which relies purely on exploitation can be greatly enhanced by completing it with the exploration queries introduced by \eqref{mixing}. 

Another option is to sample nodes that maximize the minimum over all possible labels change. That is, instead of \eqref{averaging} one can adopt utility scores of the general form
\begin{align}\label{maxmin}
U(v_i,\mathcal{L})=\min_{y_i\in \{-1,1\}} C(y_i,\mathcal{L}).
\end{align}
 Albeit intuitive, \eqref{mixing} is not as appropriate for bias reduction of more sophisticated strategies such as the ones presented in this work, since it ignores the structure of the graph, and it is somewhat aggressive in assuming that with probability $\pi^t$ the model is completely uninformative. For similar reasons, \eqref{maxmin} also does not produce satisfying results.    

In the present section, we introduce a ``softer'' bias-reduction heuristic that is better tailored to the sampling strategies at hand.
The main idea is to average over $U(v_i,\mathcal{L})$ in  \eqref{averaging} using a different set of probabilities than the ones provided by the model (cf. \eqref{posterior_approx}). Specifically, we suggest to average over label predictions that are closer to an ``non-informative'' prior early on, and gradually converge to the ones provided by the trained model as our confidence on the latter increases. Thus, instead of taking the expectation in \eqref{averaging} over $p\left( y_i | \mathbf{y}_{\mathcal{L}} \right)$, one may instead use
\begin{align}\label{alpha_approx}
\check{p}\left( y_i | \mathbf{y}_{\mathcal{L}} ; \alpha_t \right)= \alpha_t \pi(y_i) +(1-\alpha_t)p\left( y_i | \mathbf{y}_{\mathcal{L}} \right)
\end{align}
where $0\leq \alpha_t \leq 1$ is a constant that quantifies the confidence on the current estimate of the posterior.
If no prior is available, one may simply use $\pi(y_i=1)=\pi(y_i=-1)=1/2$. Intuitively pleasing results were obtained when combining \eqref{alpha_approx} with several methods. For instance, combining \eqref{alpha_approx} with our proposed TV method yields the following modified MSD utility score function   
\begin{align}
U_{\mathit{MSD}}(v_i,\mathcal{L},a_t)\propto \left[0.5a_t+ (1-a_t) \frac{(1-\mu_i^2)}{g_{ii}} \right] \frac{\|\mathbf{g}_i\|_2^2}{g_{ii}}
\end{align}
where $a_t$ tunes the sensitivity of the sampling process to the uncertainty metric $(1-\mu_i^2)$. 
As more samples become available, the confidence that the current estimate of the posterior is close to the true distribution may increase. Thus, instead of using a constant $\alpha$ throughout the sampling process, one may use a sequence $\{\alpha_t\}_{t=1}^{T}$, where $t$ is the iteration index, $T$ the total number of samples, and $a_t$ is \emph{inversely proportional} to $t$. { Finally, note that by setting $\alpha_t=1 \forall t$ the uncertainty terms vanish with MSD and TV becoming non-adaptive and equivalent to the VM and $\Sigma$-opt correspondingly.}

\section{Numerical Tests}\label{sec:simulations}

The present section includes numerical experiments carried to assess the performance of the proposed methods in terms of prediction accuracy. Specifically, the ensuing subsections present plots of accuracy $$\mathrm{Accuracy}=\frac{1}{|\mathcal{U}|}\sum_{i\in \mathcal{U}}\mathbf{1}_{\{ \hat{y}_i=y_i \}}$$
 as a function of the number of nodes sampled by the GMRF-based active learning algorithms (cf. Algorithm \ref{algorithm}). We compare the proposed methods (number of flips (FL), KL divergence, MSD, sum of TVs) with the variance minimization (VM) \cite{VM12}, $\Sigma-$optimality \cite{sigmaopt}, expected error minimization (EER) \cite{riskmin2003}, and two-step approximation method (TSA) \cite{nowak16}. Furthermore, we compare with the minimum-margin uncertainty sampling (UNC) scheme that samples the node with smallest difference between the largest soft labels, which is equivalent to using the utility function $U_{UNC}(v_i,\mathcal{L}):=-|\mu_i^{(c_1)}-\mu_i^{(c_2)}|$, where $c_1$ and $c_2$ is the most-probable and second-most probable class for node $v_i$ correspondingly. Finally, all methods are compared to the predictions that are given by a passive learning method based on random sampling. { For all graph tested the prediction accuracy remained high for a large range of $\delta\in[0.1-0.001]$ with the exact value tuned for every graph in order to maximize accuracy for passive (random) sampling.}

\subsection{Synthetic graphs}

Following \cite{nowak16}, we first considered a $10\times 10$ rectangular grid similar to the one in Fig. 1, where each node is connected to four neighboring nodes. Red dots correspond to nodes belonging to class $1$, and uncolored intersections correspond to nodes belonging to class -1. To make the classification task more challenging, the class 1 region was separated into two $3 \times 3$ squares (upper left and lower right) and additional class 1 nodes were added w.p. $0.5$ along the dividing lines. Plotted in Fig. 3 is the accuracy-vs-number of samples performance averaged over 50 Monte Carlo runs. As expected, most algorithms outperform random sampling. In addition, one observes that purely exploratory non-adaptive methods (VM and $\Sigma-$optimality) enjoy relatively high accuracy for a small number of samples, but are eventually surpassed by adaptive methods. It can also be observed that the novel TV method with $a_t=t^{-1/2}$ performs equally well to the state-of-the-art TSA method. Interestingly, it does so while using a much simpler criterion that avoids model retraining, and therefore requires significantly shorter runtime. { Note finally that the performance of ERR is poor because the sampler easily becomes ``trapped'' in one of the two class 1 regions, and does not explore the graph. }

{ The purpose of the experiment in Fig. 1 was to simulate problems where a complex label distribution appears on a simple uniform graph (e.g., image segmentation). To simulate more structured graphs, we generated a 1000-node network using the Lancichinetti--Fortunato--Radicchi (LFR) method \cite{lancichinetti2008benchmark}. The LFR algorithm is widely used to generate benchmark graphs that resemble real world networks by exhibiting community structure and degree distributions that follow the power law. Figure 2 reveals the sparsity pattern of the adjacency matrix of the LFR graph that was used, while the 3 clearly visible clusters correspond to groups of nodes in the same class, that is
	\begin{equation*}
	{y}_i=
	\left\{ \begin{array}{cc}
	1,~&~i\in [1,250]\\
	2,~&~i\in [251,600]\\
	3,~&~i\in [601,1000]
	\end{array} \right. 
	\end{equation*}
Note that, unlike the one in Fig. 1, the graph used here is characterized by a community structure that matches the nodes labels. This is a highly favorable scenario for the non-adaptive VM and $\Sigma$-opt approaches that rely solely on the graph structure. Indeed, as seen in Fig. 4, VM and $\Sigma$-opt quickly reach 90$\%$ accuracy by selecting 5 most influential samples. Nevertheless, between 5 and 10 samples our proposed MSD and TV adaptive methods enjoy superior accuracy before converging to 100$\%$ accuracy.   
}
\begin{figure*}[ht] \label{ synthetic} 
	\begin{minipage}{0.5\linewidth}
		\centering
		\includegraphics[width=0.65\linewidth]{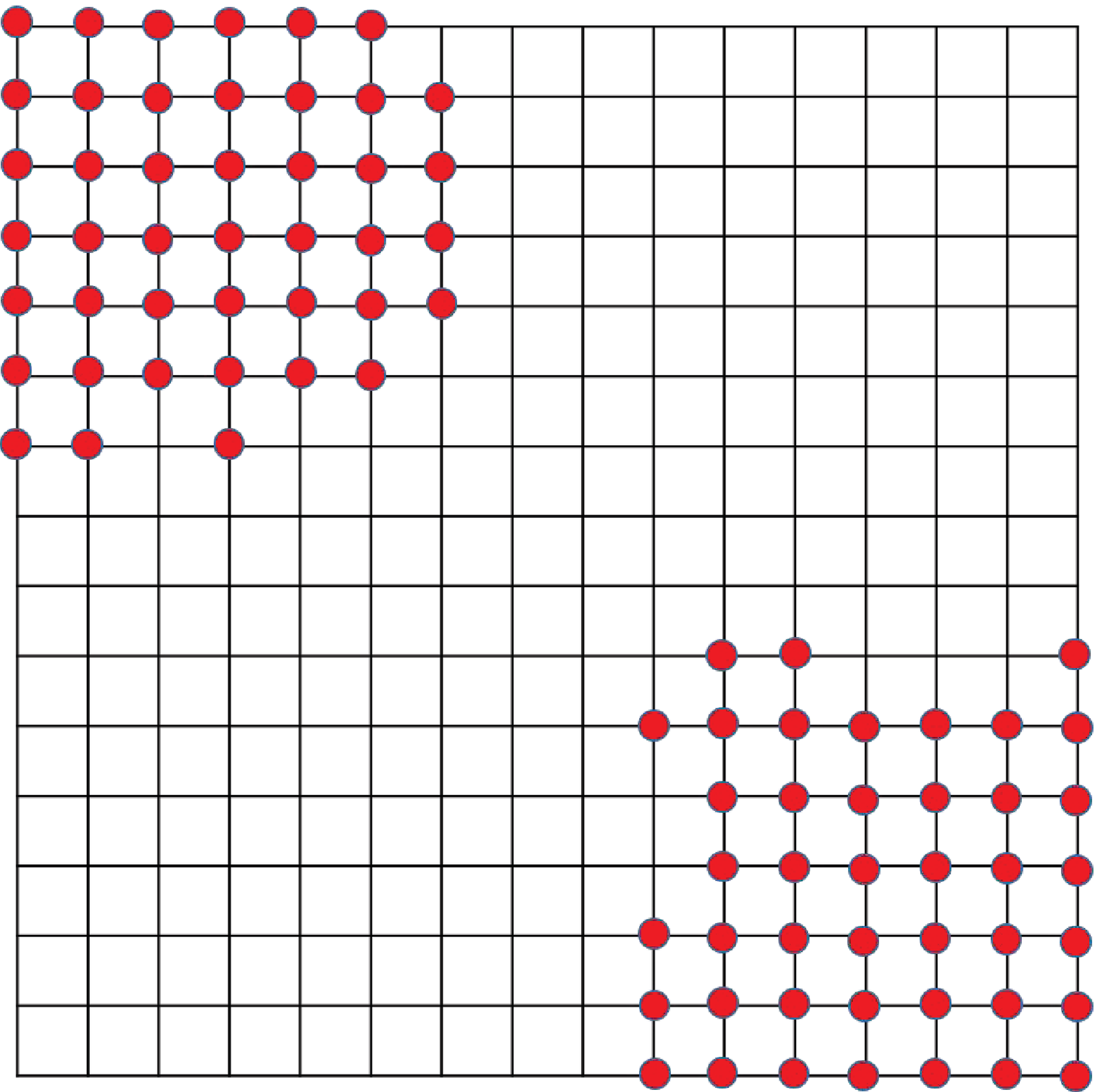} 
		\caption{Rectangular grid synthetic graph with two separate class 1 regions.} 
	\end{minipage} 
	\begin{minipage}{0.5\linewidth}
		\centering
		\includegraphics[width=0.85\linewidth]{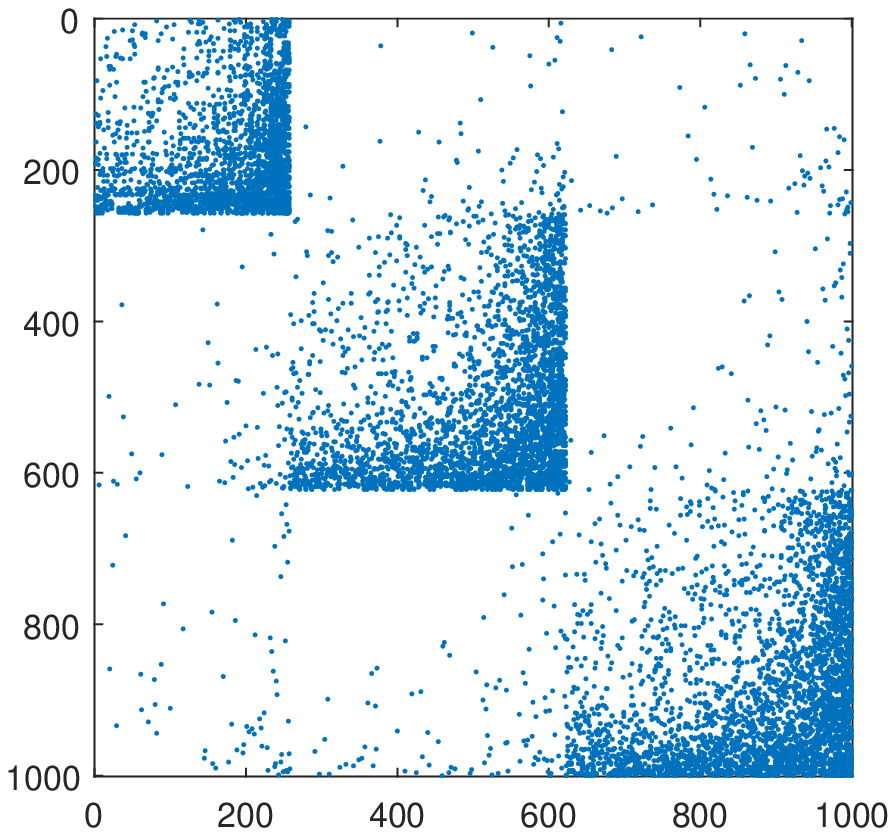}\hspace{10cm} 
		\vspace{-0.7cm}
		\caption{Adjacency matrix of  LFR graph with 1,000 nodes and 3 classes. } 
	\end{minipage} 
	\begin{minipage}{0.5\linewidth}
		\centering
		\includegraphics[width=0.95\linewidth]{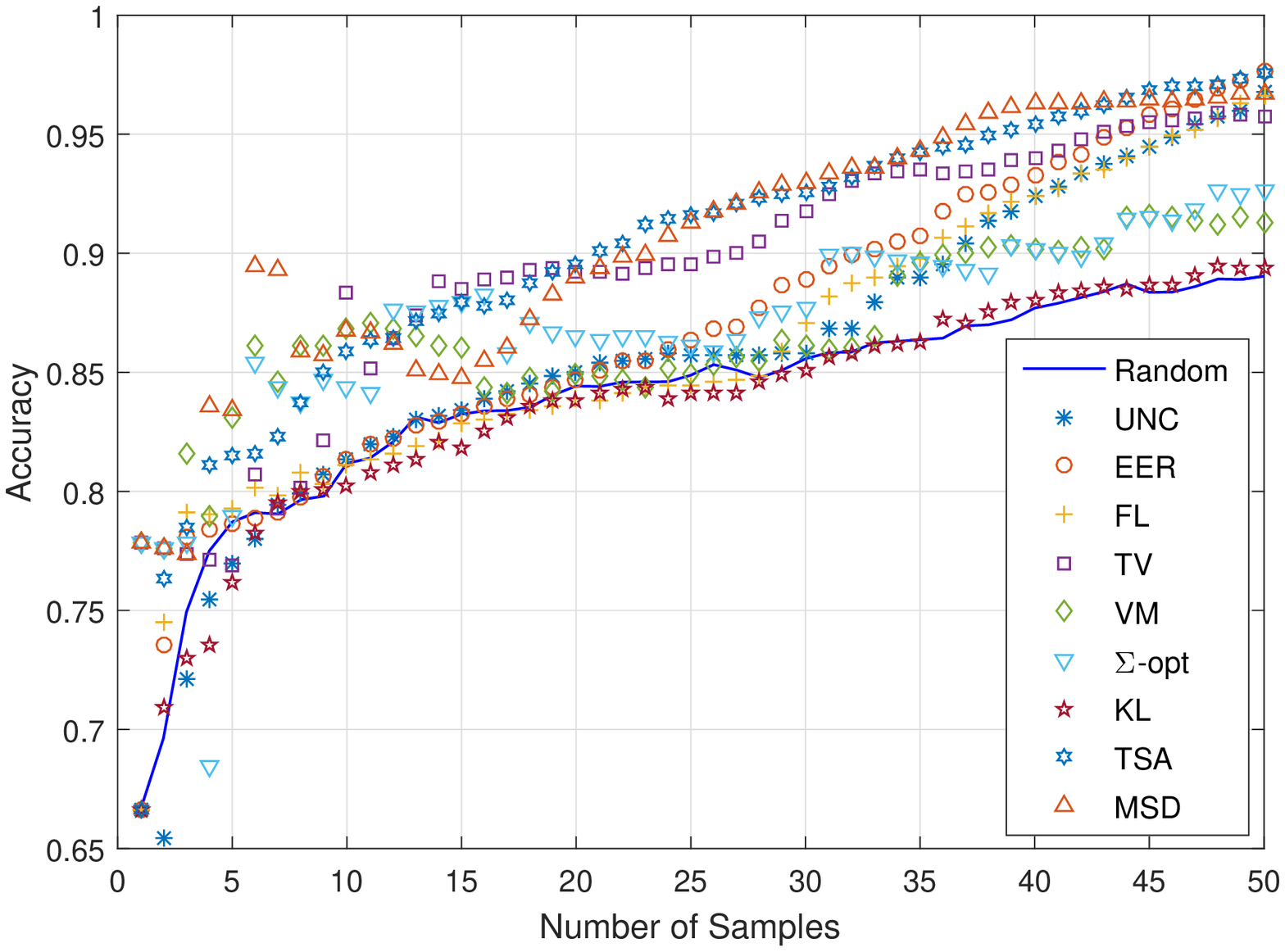} 
		\caption{Test results for synthetic grid in Fig. 1 . } 
	\end{minipage}
	\begin{minipage}{0.5\linewidth}
		\centering
		\includegraphics[width=0.95\linewidth]{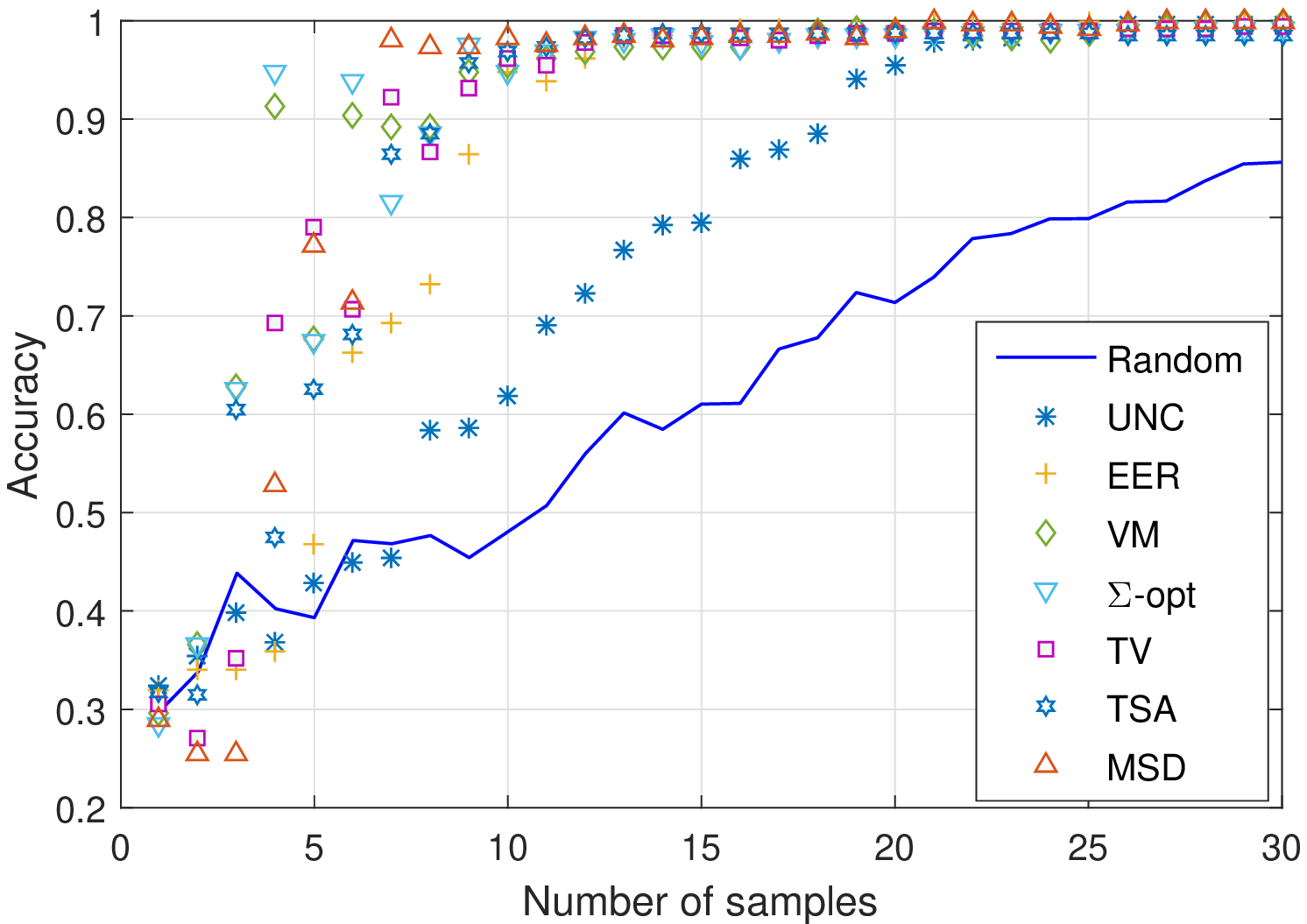} 
		\caption{Test results for synthetic LFR graph in Fig. 2.} 
	\end{minipage} 
\end{figure*}

%

			
%
%
%
			\subsection{Similarity graphs from real datasets}

			Real binary classification datasets taken from the UC Irvine Machine Learning Repository \cite{uci} and the LibSVM webpage \cite{libsvm} were used for further testing of the proposed methods. First, each entry of the feature vectors was normalized to lie between -1 and 1. Then, a graph was constructed using the Pearson correlations among pairs of normalized feature vectors as weights of the adjacency matrix $\mathbf{W}$; thresholding was also applied to negative and small weights leading to sparse adjacency matrices. It was observed that sparsification generally improves the prediction accuracy, while also reducing the computational burden. In the presented experiments, thresholds were tuned until one of the methods achieved the highest possible classification accuracy. 
			
			Having constructed the graphs, the proposed expected model change sampling schemes were compared with UNC, TSA, EER, VM and $\Sigma-$optimality on seven real datasets listed in Table II; in the latter, ``baseline accuracy" refers to the proportion of the largest class in each dataset, and thus the highest accuracy that can be achieved by naively assuming that all labels belong to the majority class. Plotted in Figs. 5 to 10 are the results of the numerical tests, where it is seen that the performance of the proposed low-complexity TV- and MSD-based scheme is comparable or superior to that of competing alternatives. The confidence parameter was set to $a_t=1/\sqrt{t}$ for the smaller datasets, where only few data were sampled, and the model was expected to be less accurate, whereas for the larger ones it was set to $a_t=0$. 		    
           \begin{table}[t]\label{tab:table_1}
           	\caption{ Dataset list }
           	\begin{center}
           		\begin{tabular} {|c|c|c|}
           			\hline
           			Dataset & \# of nodes & Baseline Accuracy \\
           			\hline 
           			Coloncancer  & 62  &  0.64 \\
           			\hline 
           			Ionosphere &  351  & 0.64 \\
           			\hline
           			Leukemia &  70 & 0.65 \\
           			\hline
           			Australian & 690 & 0.55\\
           			\hline
           			Parkinsons & 191 & 0.75\\
           			\hline
           			Ecoli & 326 & 0.57 \\
           			\hline
           		\end{tabular}
           	\end{center}
           \end{table}
      \begin{figure*}[ht] \label{data} 
      	\begin{minipage}{0.5\linewidth}
      		\centering
      		\includegraphics[width=0.9\linewidth]{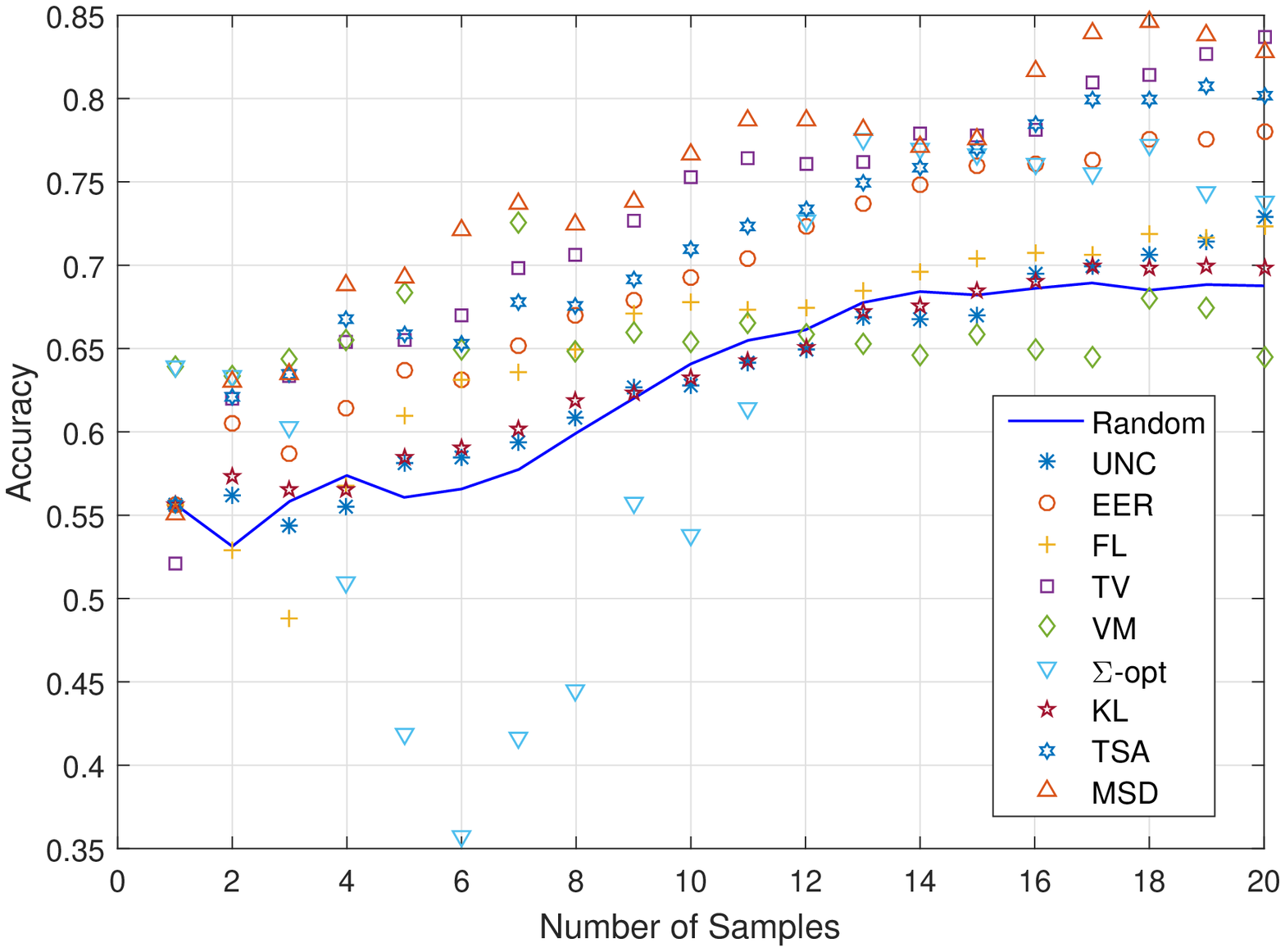} 
      		\caption{Coloncancer dataset.} 
      	\end{minipage} 
      	\begin{minipage}{0.5\linewidth}
      		\centering
      		\includegraphics[width=0.9\linewidth]{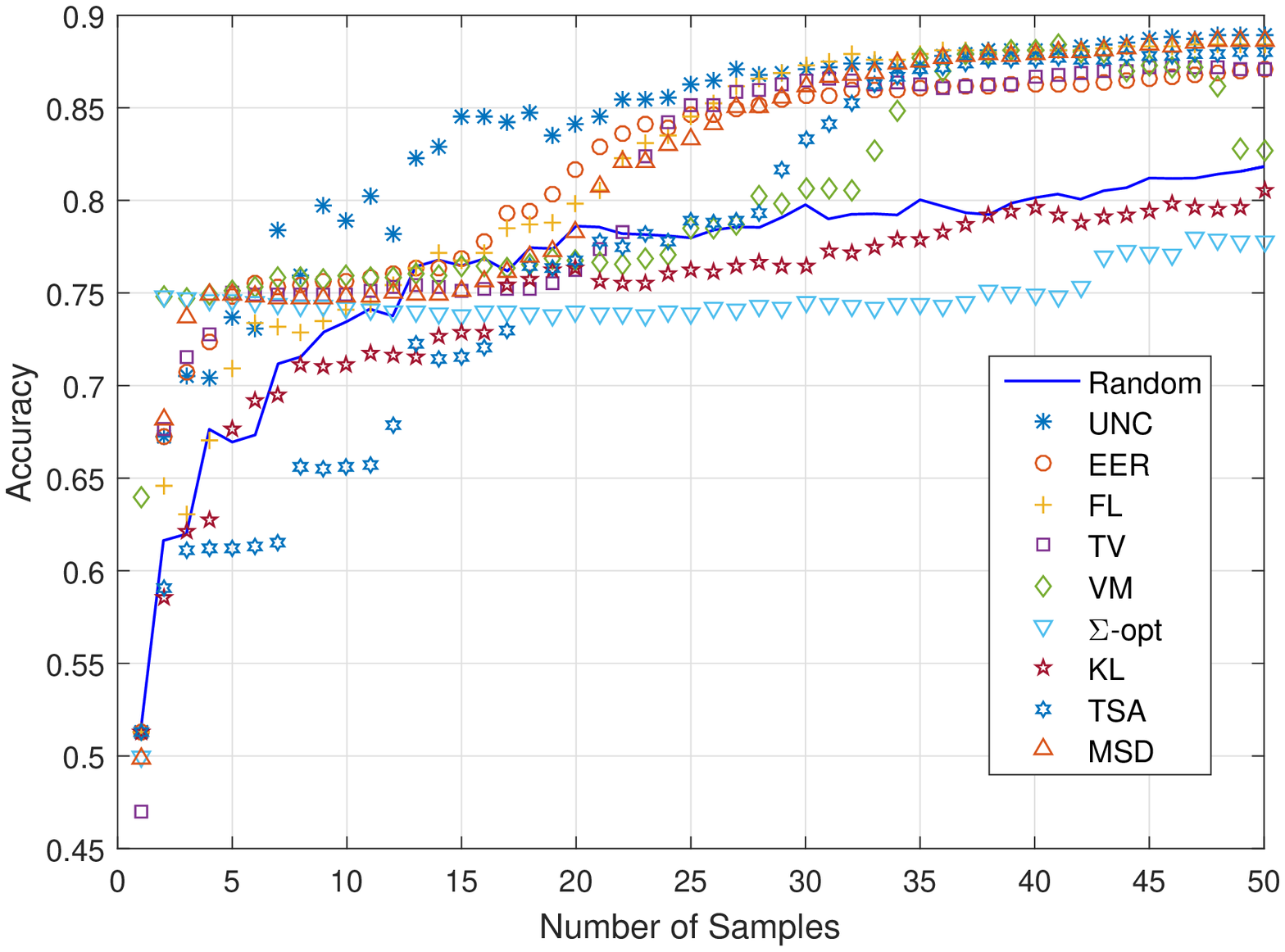} 
      		\caption{Ionosphere dataset.} 
      	\end{minipage} 
      	\begin{minipage}{0.5\linewidth}
      		\centering
      		\includegraphics[width=0.9\linewidth]{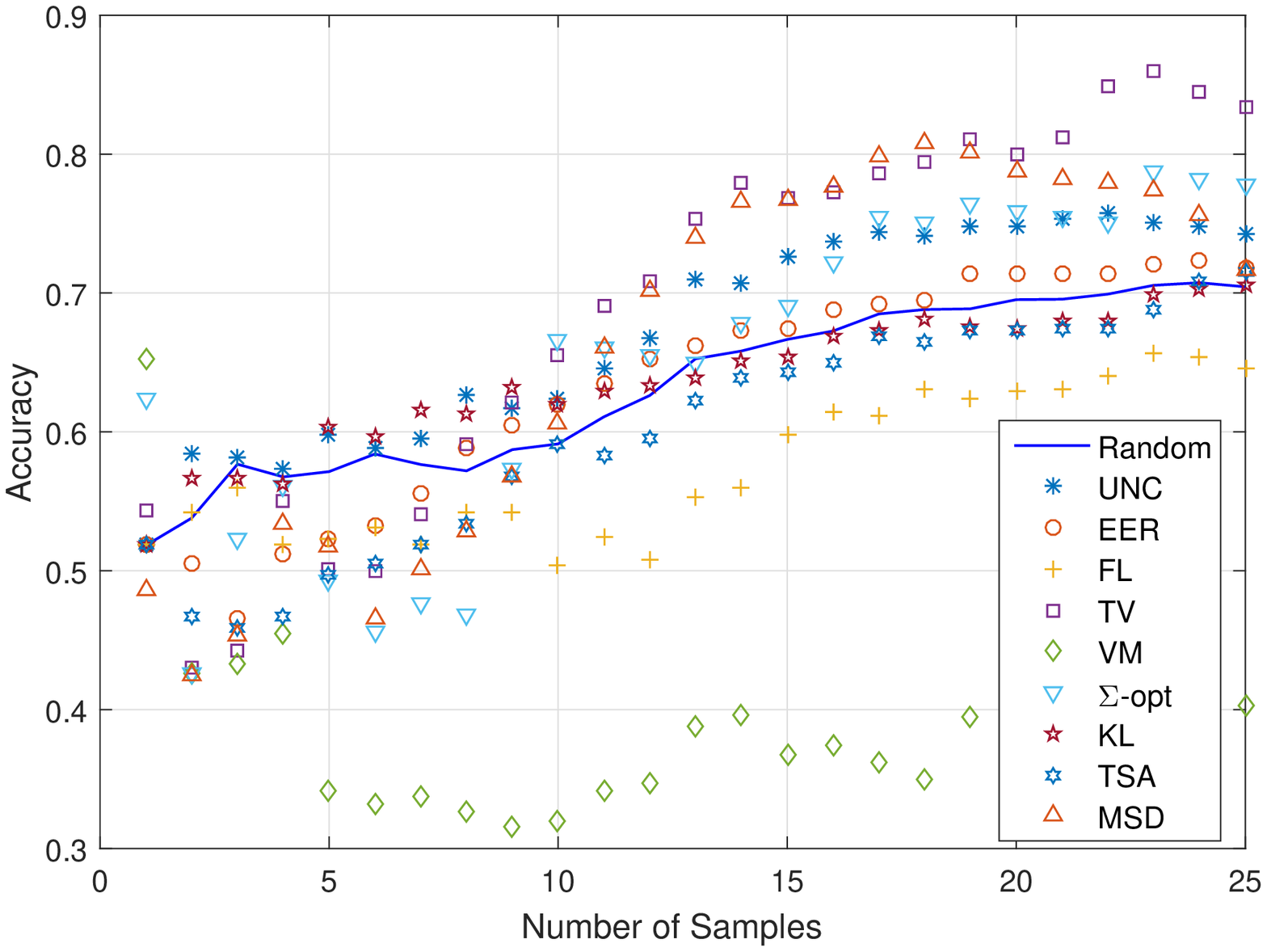} 
      		\caption{Leukemia dataset. } 
      	\end{minipage}
      	\begin{minipage}{0.5\linewidth}
      		\centering
      		\includegraphics[width=0.9\linewidth]{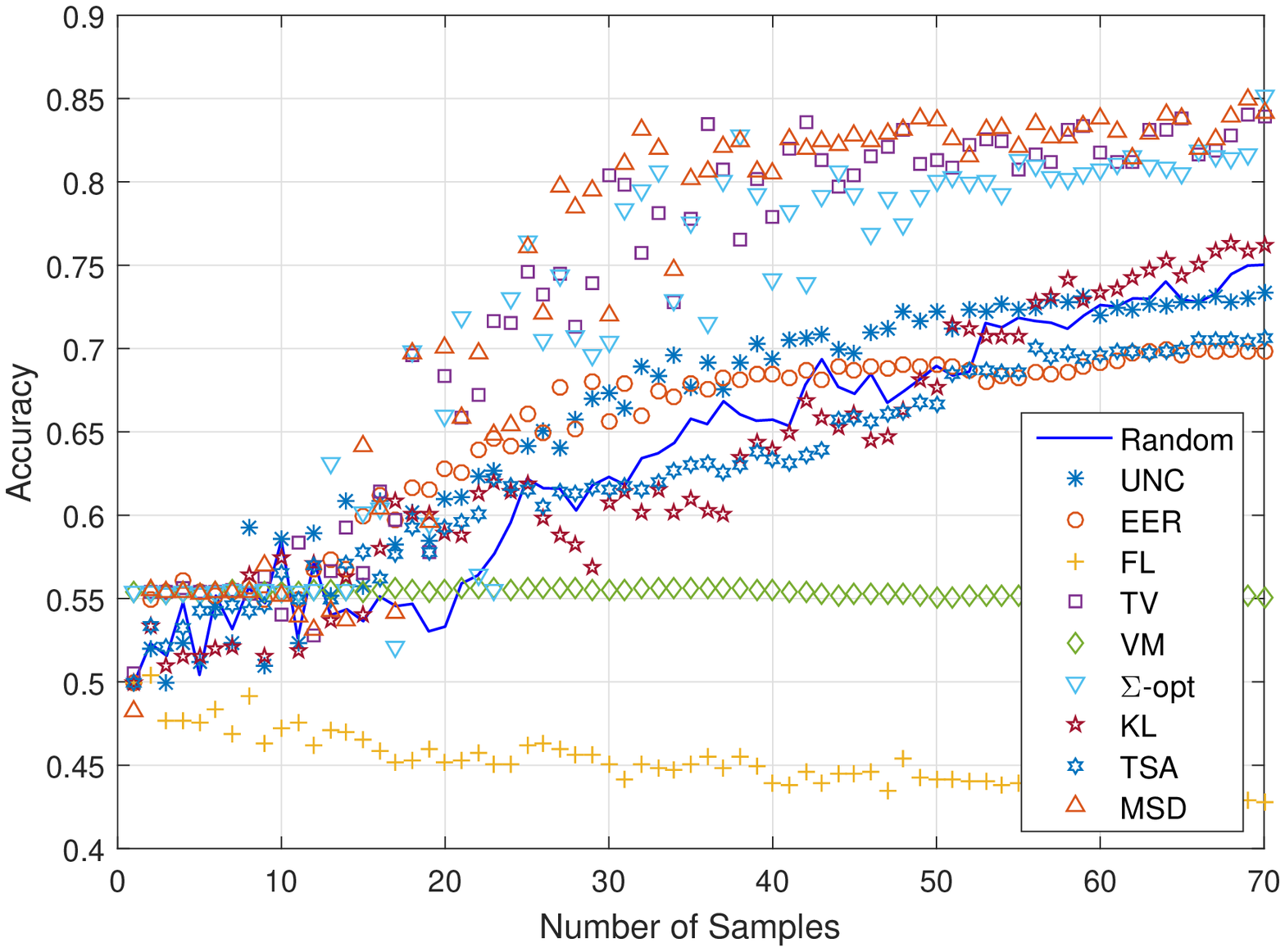} 
      		\caption{Australian dataset.} 
      	\end{minipage} 
      	\begin{minipage}{0.5\linewidth}
      		\centering
      		\includegraphics[width=0.9\linewidth]{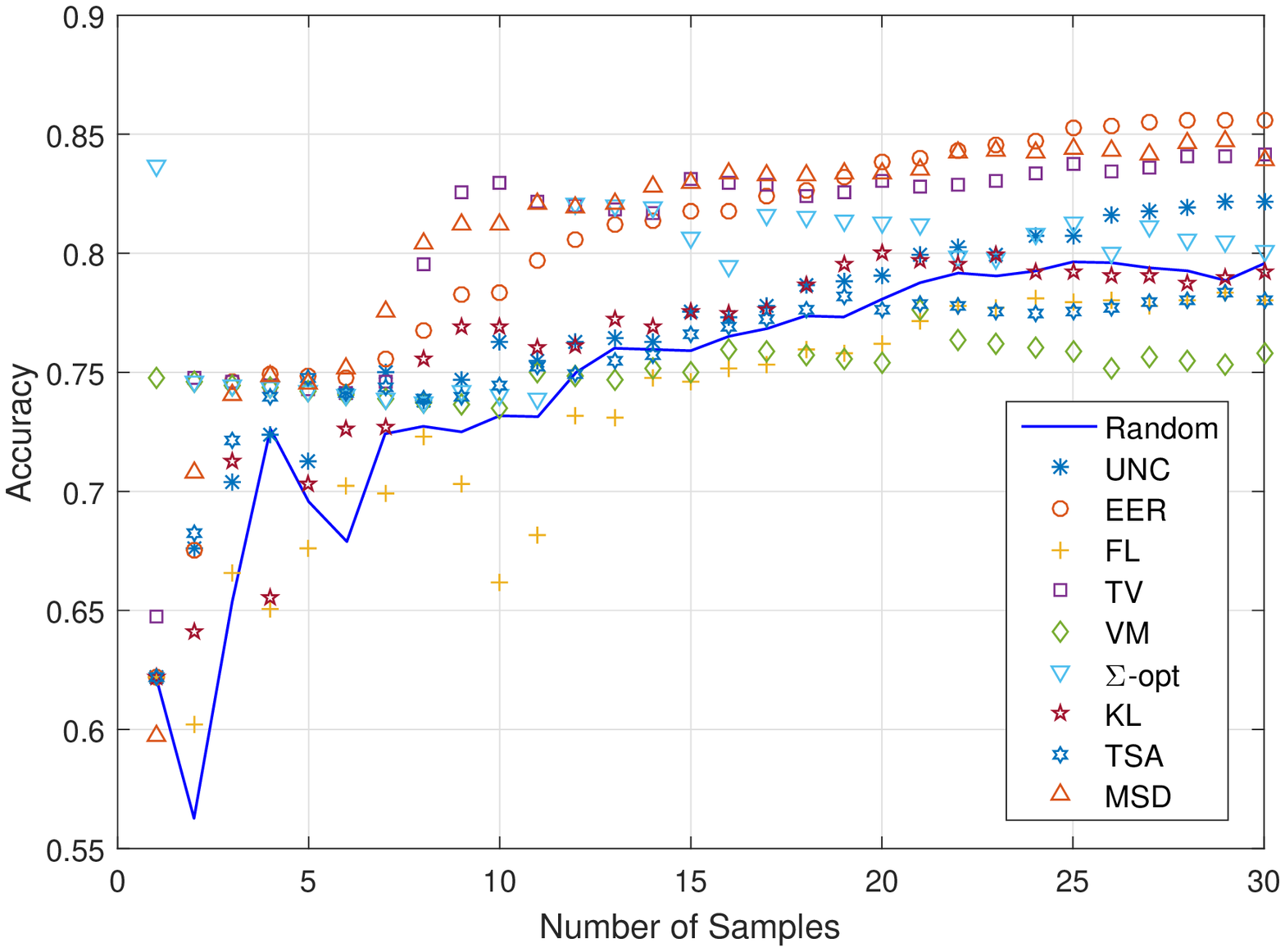} 
      		\caption{Parkinsons dataset.} 
      	\end{minipage} 
      	\begin{minipage}{0.5\linewidth}
      		\centering
      		\includegraphics[width=0.9\linewidth]{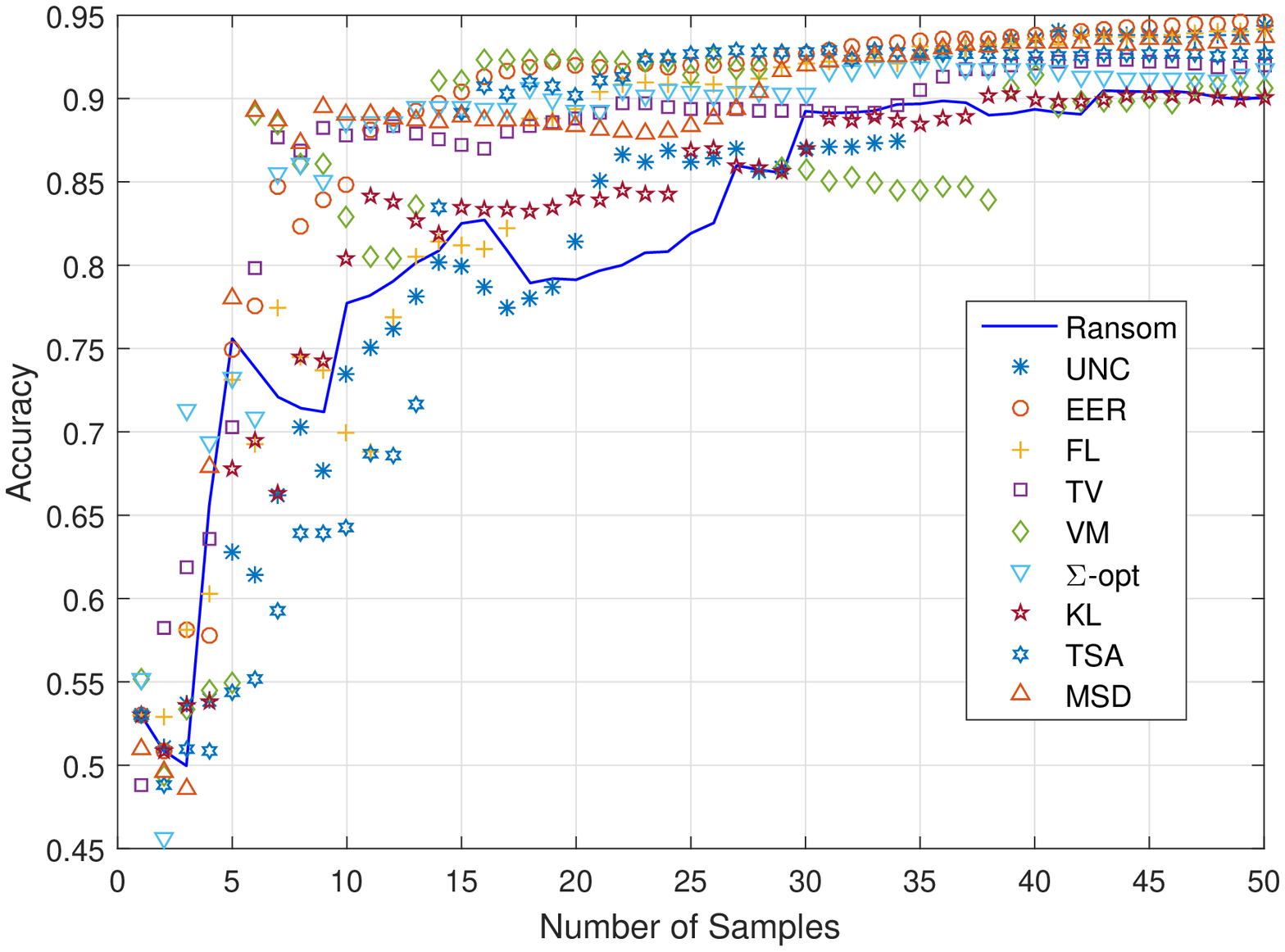} 
      		\caption{Ecoli dataset.} 
      	\end{minipage}
      \end{figure*}

	{		
		\subsection{Real graphs}
		Experiments were also performed on real labeled graphs. Specifically, the CORA and CITESEER \cite{cora} citation networks with 2708 and 3312 nodes correspondingly were used; similarly to \cite{sigmaopt}, we isolated the largest connected components. In citation networks, each node corresponds to a scientific publication and is linked only with cited or citing papers. Nodal labels correspond to the scientific field that each paper belongs to (6 classes for CITESEER and 7 for CORA). The benchmark political-blog network \cite{polblog} with 1490 nodes and two classes was also used. The confidence sequence $\alpha_t=t^{-1/2}$ was used for all graphs, with $\delta = 0.005$ similarly to [11]. The results of the experiments are depicted in Figs. 11-13 and demonstrate the effectiveness of the proposed MSD and TV algorithms on these social graphs. For the CORA network, TV achieves state of the art performance equal to EER, TSA and $\Sigma$-opt, while for the CITESEER network its accuracy slightly surpasses that of competing methods. For the political-blogs network, non-adaptive TV and $\Sigma$-opt methods perform poorly,  while the proposed MSD method performs at least as good as the significantly more complex TSA. The bar plot in Fig. 14 depicts the relative runtimes of different adaptive methods. Observe that MSD and TV are two orders of magnitude faster than EER and TSA for the larger multilabel citation graphs, and one order of magnitude faster for the smallest binary-labeled political blogs network.
	}
	      \begin{figure}[h] \label{nets} 
	      	\begin{minipage}{1\linewidth}
	      		\centering
	      		\includegraphics[width=0.95\linewidth]{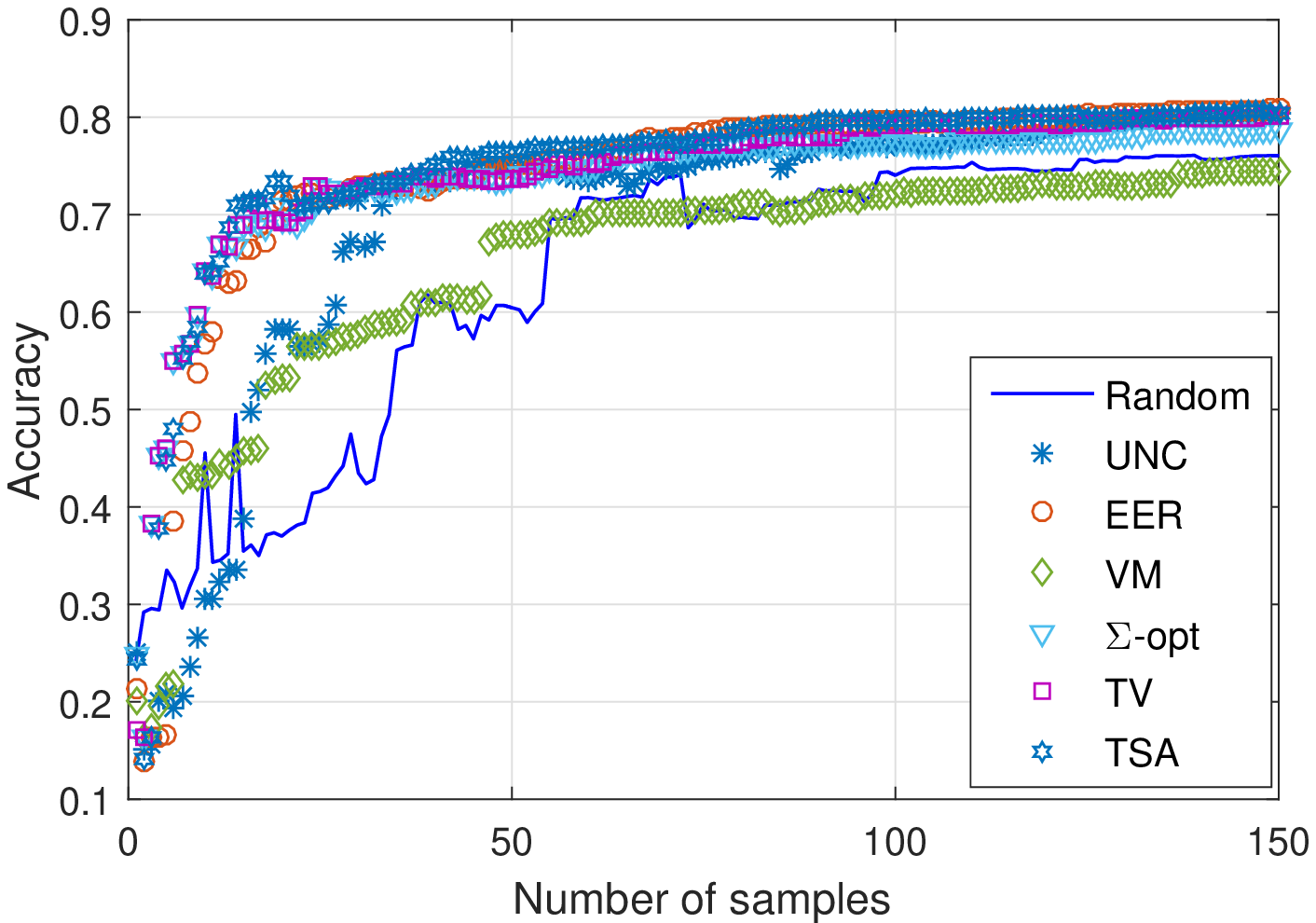} 
	      		\caption{CORA citation network.} 
	      	\end{minipage} 
	      	\begin{minipage}{1\linewidth}
	      		\centering
	      		\includegraphics[width=0.95\linewidth]{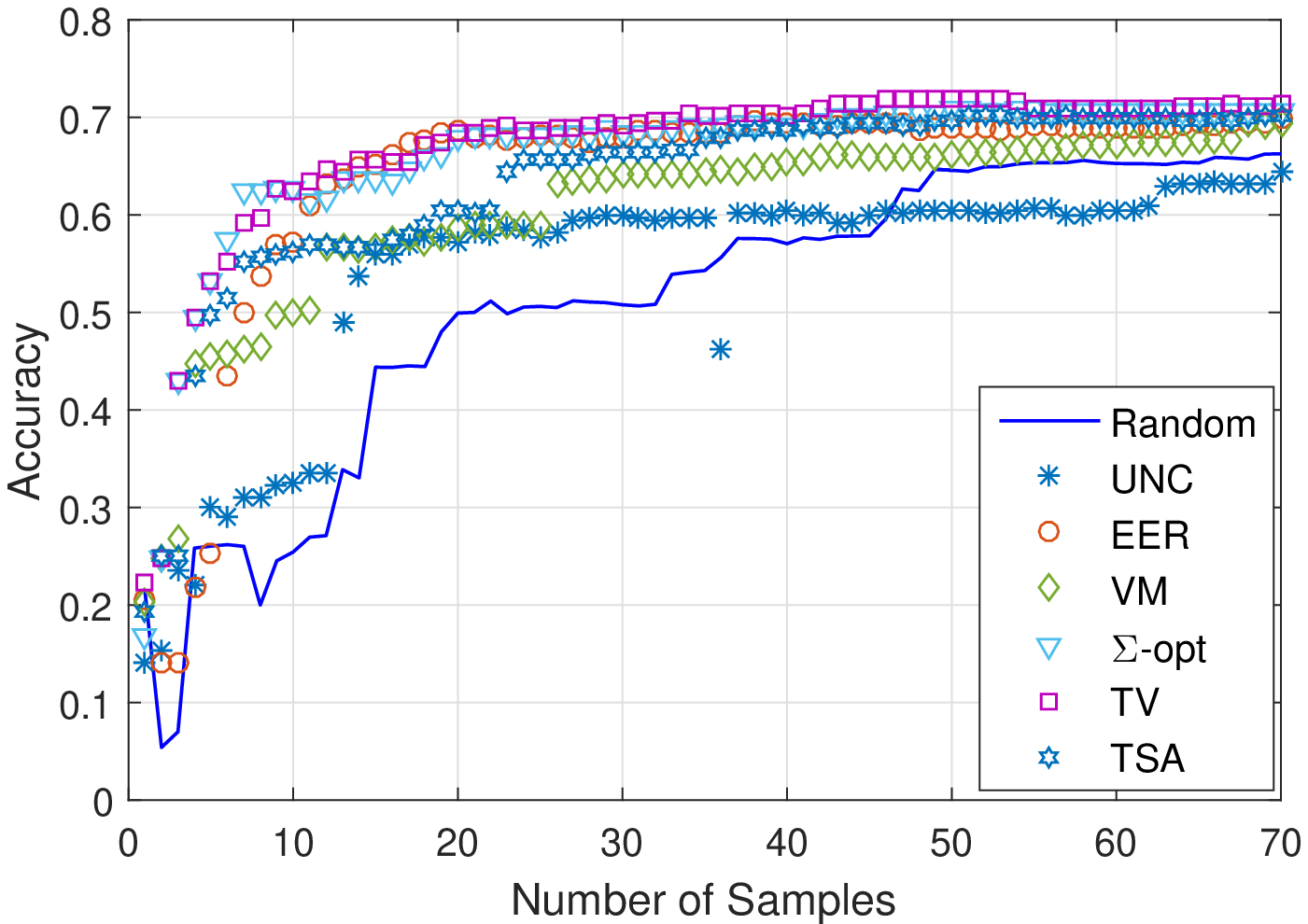} 
	      		\caption{CITESEER citation network.} 
	      	\end{minipage} 
	      	\begin{minipage}{1\linewidth}
	      		\centering
	      		\includegraphics[width=0.95\linewidth]{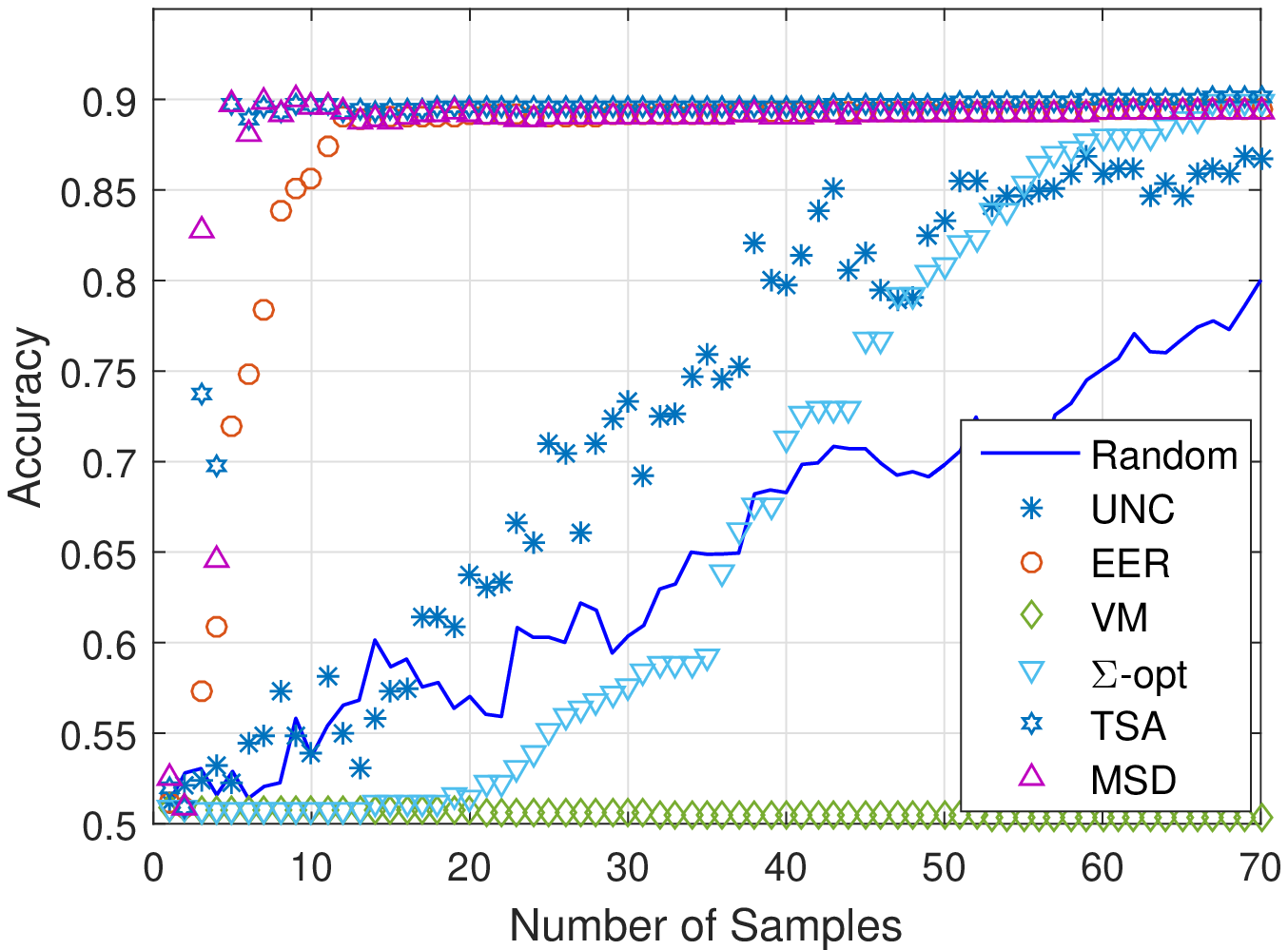} 
	      		\caption{Political blogs network. } 
	      	\end{minipage}
	      \end{figure}
	
	\begin{figure}[ht] \label{ runtime} 
			\centering
			\includegraphics[width=1\linewidth]{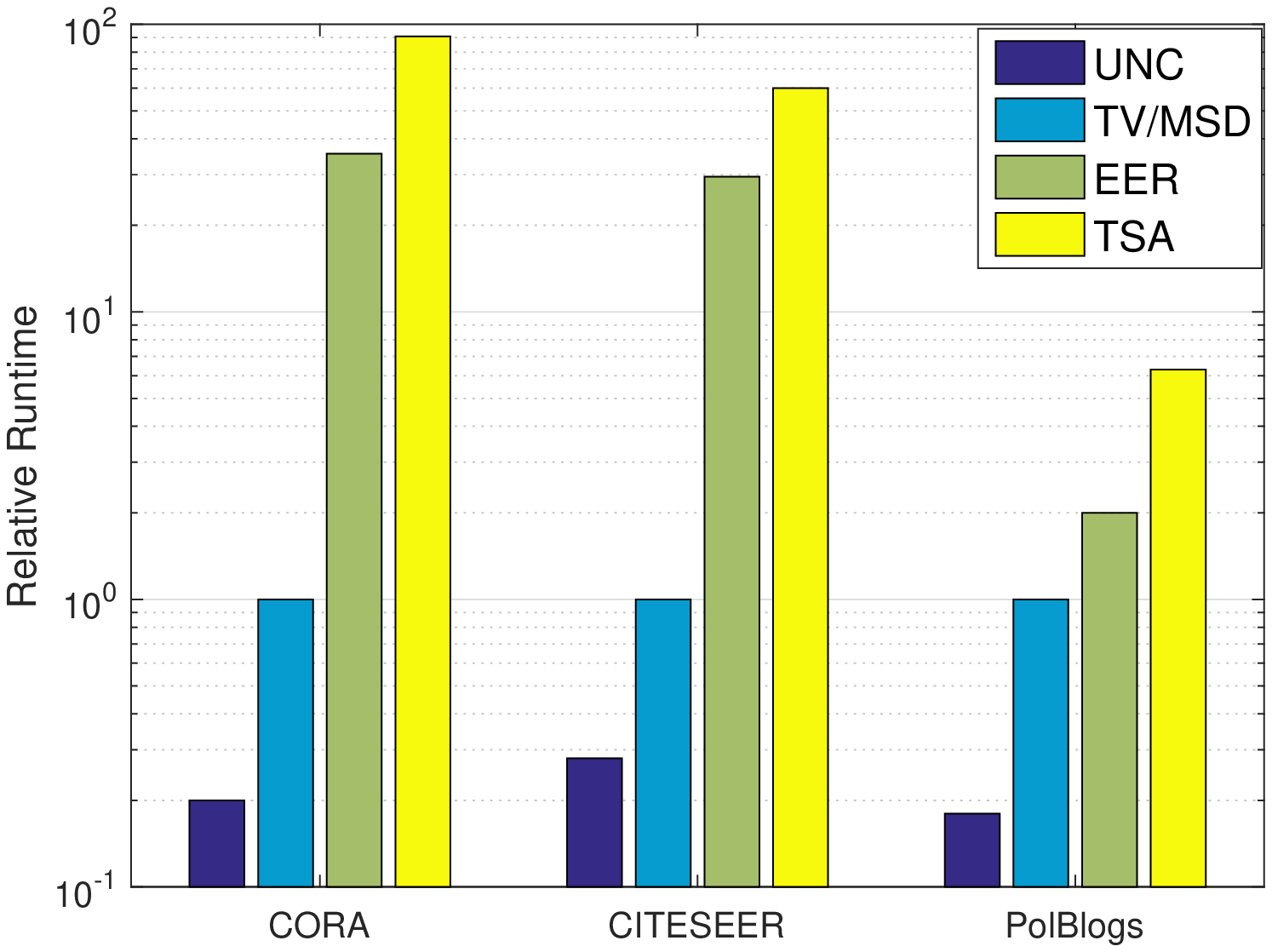} 
			\caption{Relative runtime of different adaptive methods for experiments on real social graphs.} 
	\end{figure}

      					
\section{Conclusions and Future Directions}\label{sec:conclusions}
This paper unified existing and developed novel utility functions for data-adaptive graph-cognizant active classification using GMRFs.
	 These utility functions rely on metrics that capture expected changes in GMRF models. Specifically, the proposed samplers query the node that is expected to inflict the largest change on the model. Towards this direction, several measures of expected model change were introduced, sharing similarities and connections with existing methods such as uncertainty sampling, variance minimization, and sampling based on the $\Sigma-$optimality criterion. A simple yet effective heuristic was also introduced for increasing the exploration capabilities and reducing bias of the proposed methods, by taking into account the confidence on the model label predictions. Numerical tests using synthetic and real data confirm that the proposed methods achieve accuracy that is comparable or superior to state of the art at smaller runtime.
	 
	 Future research directions will focus on developing even more efficient adaptive sampling schemes for graphs by finding the sweet spot of how a given graph structure attains the desirable exploration versus exploitation trade-off. Furthermore, our research agenda includes developing adaptive sampling methods tailored for Markov-chain-Monte-Carlo-based and random-walk-based inference on graphs.

\appendix

\textbf{A1}. Since $\mathbf{C}^{-1}=\mathbf{L}$ and upon partitioning the two matrices according to labeled and unlabeled nodes, we have
\begin{equation}\label{matrix_equation}
\left[\begin{array}{cc}
\mathbf{L}_{\mathcal{UU}} & \mathbf{L}_{\mathcal{UL}} \\
\mathbf{L}_{\mathcal{LU}} & \mathbf{L}_{\mathcal{LL}} \end{array} \right]
\left[\begin{array}{cc}
\mathbf{C}_{\mathcal{UU}} & \mathbf{C}_{\mathcal{UL}} \\
\mathbf{C}_{\mathcal{LU}} & \mathbf{C}_{\mathcal{LL}} \end{array} \right]=
\left[\begin{array}{cc}
\mathbf{I}_{|\mathcal{U}|} & \mathbf{0} \\
\mathbf{0} & \mathbf{I}_{|\mathcal{L}|} \end{array} \right]
\end{equation}
which gives rise to four matrix equations. Specifically, the equation that corresponds to the upper right part of \eqref{matrix_equation} is
\begin{equation}\label{matrix_equation2}
\mathbf{L}_{\mathcal{UU}}\mathbf{C}_{\mathcal{UL}} + \mathbf{L}_{\mathcal{UL}}\mathbf{C}_{\mathcal{LL}}=\mathbf{0} .
\end{equation}
Multiplying \eqref{matrix_equation2} from the left by $\mathbf{L}_{\mathcal{UU}}^{-1}$ and from the right by $\mathbf{C}_{\mathcal{LL}}^{-1}$ yields
$
\mathbf{C}_{\mathcal{UL}}\mathbf{C}_{\mathcal{LL}}^{-1}=- \mathbf{L}_{\mathcal{UU}}^{-1}\mathbf{L}_{\mathcal{UL}},
$
which verifies \eqref{cond_exp}.

\textbf{A2}. Let $\mathbf{x}_1\sim \mathcal{N}(\mathbf{m}_1,\mathbf{C})$ and $\mathbf{x}_2\sim \mathcal{N}(\mathbf{m}_2,\mathbf{C})$, and assume that $\mathbf{x}_1$ and $\mathbf{x}_2$ are uncorrelated. Then,
\begin{align}\nonumber
\mathrm{MSD}(\mathbf{x}_1,\mathbf{x}_2)&:=\mathbb{E}\left[\|\mathbf{x}_1-\mathbf{x}_2\|_2^2\right]\\\label{0}
&=\mathbb{E}\left[\|\mathbf{x}_1\|_2^2+\|\mathbf{x}_2\|_2^2-2\mathbf{x}_1^T\mathbf{x}_2\right]
\end{align}
where
\begin{align}\nonumber
\mathbb{E}\left[\|\mathbf{x}_1\|_2^2\right]&=\mathbb{E}\left[\|(\mathbf{x}_1-\mathbf{m}_1)+\mathbf{m}_1\|_2^2\right]\\\nonumber
&=\mathbb{E}\left[\|\mathbf{x}_1-\mathbf{m}_1\|_2^2\right]+2\mathbb{E}\left[(\mathbf{x}_1-\mathbf{m}_1)^T\mathbf{m}_1\right]+\|\mathbf{m}_1\|_2^2\\ \label{1}
&=\mathrm{tr}\left(\mathbf{C}\right) + \|\mathbf{m}_1\|_2^2
\end{align}
and similarly for $\mathbb{E}\left[\|\mathbf{x}_2\|_2^2\right]$. Finally, note that 
\begin{align}\nonumber
\mathbb{E}\left[(\mathbf{x}_1-\mathbf{m}_1)^T(\mathbf{x}_2-\mathbf{m}_2)\right]
&=\mathbb{E}\big[\mathbf{x}_1^T\mathbf{x}_2-\mathbf{x}_1^T\mathbf{m}_2\\\nonumber
&-\mathbf{m}_1^T\mathbf{x}_2+\mathbf{m}_1^T\mathbf{m}_2\big]\\ \label{intermediate}
&=\mathbb{E}\left[\mathbf{x}_1^T\mathbf{x}_2\right]-\mathbf{m}_1^T\mathbf{m}_2
\end{align}
and since  $\mathbf{x}_1$ and $\mathbf{x_2}$ are uncorrelated it follows that \eqref{intermediate} equals to $0$; hence,
\begin{equation}\label{2}
\mathbb{E}\left[\mathbf{x}_1^T\mathbf{x}_2\right]=\mathbf{m}_1^T\mathbf{m}_2.
\end{equation} 
Substituting \eqref{1} and \eqref{2} into \eqref{0} yields
\begin{align*}
\mathrm{MSD}(\mathbf{x}_1,\mathbf{x}_2)&=2\mathrm{tr}(\mathbf{C})+\|\mathbf{m}_1\|_2^2+\|\mathbf{m}_2\|_2^2-2\mathbf{m}_1^T\mathbf{m}_2\\
&=2\mathrm{tr}(\mathbf{C})+\|\mathbf{m}_1-\mathbf{m}_2\|_2^2
\end{align*}
which implies that the MSD between two Gaussian fields with the same covariance matrix is proportional to the Euclidean norm of the difference of their means.  

\bibliographystyle{IEEEbib}
\bibliography{ref_censoring}
\end{document}